\documentclass{article}

\usepackage{microtype}
\usepackage{graphicx}
\usepackage{multirow}
\usepackage{subfigure}
\usepackage{booktabs} 

\usepackage{hyperref}

\usepackage[accepted]{icml2024}

\usepackage{amsmath}
\usepackage{amssymb}
\usepackage{mathtools}
\usepackage{amsthm}

\usepackage[capitalize,noabbrev]{cleveref}
\usepackage{xcolor}
\definecolor{darkgreen}{RGB}{0,100,0}
\definecolor{mediumgreen}{RGB}{0,128,0}
\definecolor{lightgreen}{RGB}{0,255,0}
\definecolor{forestgreen}{RGB}{34,139,34}
\definecolor{limegreen}{RGB}{50,205,50}

\newcommand{\update}[1]{{\textcolor{black}{#1}}}

\newcommand{\lms}[1]{{\textcolor{black}{#1}}}
\newcommand{\camera}[1]{{\textcolor{black}{#1}}}

\theoremstyle{plain}

\theoremstyle{definition}

\theoremstyle{remark}

\usepackage{amssymb}

\usepackage{hyperref} 
\usepackage{multirow} 
\usepackage{booktabs} 

\usepackage[textsize=tiny]{todonotes}
\usepackage{colortbl}
\usepackage{xcolor} 
\usepackage{csquotes}

\definecolor{mygreen}{RGB}{31, 177, 79} 

\begin{document}

\twocolumn[

\icmltitle{TimeSiam: A Pre-Training Framework for Siamese Time-Series Modeling}



\icmlsetsymbol{equal}{*}

\begin{icmlauthorlist}
\icmlauthor{Jiaxiang Dong}{equal,software}
\icmlauthor{Haixu Wu}{equal,software}
\icmlauthor{Yuxuan Wang}{software}
\icmlauthor{Yunzhong Qiu}{software}
\icmlauthor{Li Zhang}{software}
\icmlauthor{Jianmin Wang}{software}
\icmlauthor{Mingsheng Long}{software}
\end{icmlauthorlist}



\icmlaffiliation{software}{School of Software, BNRist, Tsinghua University. Jiaxiang Dong $<$djx20@mails.tsinghua.edu.cn$>$. Haixu Wu $<$wuhx23@mails.tsinghua.edu.cn$>$}
\icmlcorrespondingauthor{Mingsheng Long}{mingsheng@tsinghua.edu.cn}

\icmlkeywords{Machine Learning, ICML}

\vskip 0.3in
]



\printAffiliationsAndNotice{\icmlEqualContribution} 

\begin{abstract}

Time series pre-training has recently garnered \lms{wide} attention for its potential to reduce labeling expenses and benefit various downstream tasks. Prior methods are mainly based on pre-training techniques well-acknowledged in vision or language, such as masked modeling and contrastive learning. However, randomly masking time series or calculating series-wise similarity will distort or neglect inherent temporal correlations crucial in time series data. To emphasize temporal correlation modeling, this paper proposes TimeSiam as a simple but effective self-supervised pre-training framework \lms{for Time series} based on Siamese networks. Concretely, TimeSiam pre-trains Siamese encoders to capture intrinsic temporal correlations between randomly sampled past and current subseries. With a simple data augmentation method (e.g.~masking), TimeSiam can benefit from diverse augmented subseries and learn internal time-dependent representations through a past-to-current reconstruction. Moreover, learnable lineage embeddings are also introduced to distinguish temporal distance between sampled series and further foster the learning of diverse temporal correlations. TimeSiam consistently outperforms extensive advanced pre-training baselines, demonstrating superior forecasting and classification capabilities across 13 standard benchmarks in both \lms{intra-} and cross-domain scenarios. Code is available at \href{https://github.com/thuml/TimeSiam}{https://github.com/thuml/TimeSiam}.

\end{abstract}

\begin{figure}[t]
\begin{center}
    \centerline{\includegraphics[width=0.5\textwidth]{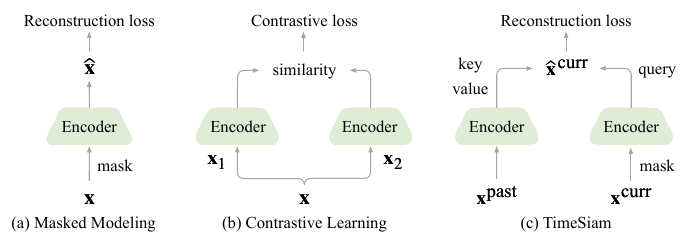}}
    \vspace{-10pt}
    \caption{{Comparison on time series pre-training \lms{frameworks}}. {(a)} Masked modeling: reconstruct the masked series. {(b)} Contrastive learning: \update{Repulse different series (negative pairs) while attracting two augmentations from the same series (positive pairs).} {(c)} TimeSiam: reconstruct masked current series $\mathbf{x}^{\text{curr}}$ from randomly sampled past observation $\mathbf{x}^{\text{past}}$.}
    \label{fig:mask_schedule}
    \vspace{-35pt}
\end{center}
\end{figure}

\vspace{-15pt}
\section{Introduction}

Time series, a critical form of real-world data, finds wide applications in various domains, including energy, traffic, economics, weather, medicine, etc \cite{wu2021-autoformer, Zhang2022-TF-C, wu2023corrformer}. In the real world, an enormous volume of time series data is incrementally collected through the Internet of Things (IoT) from industrial sensors and wearable devices. To utilize these large amounts of data, time series self-supervised pre-training has recently gained significant attention, which can extract valuable knowledge from unlabeled data and further boost the performance of various downstream tasks \cite{dong2023simmtm}. This paper focuses on this promising area and proposes a novel and practical self-supervised pre-training method for time series.

Previous pre-training methods can be roughly categorized into the following two paradigms. As presented in Figure \ref{fig:mask_schedule}, the first one, named masked modeling, enables representation learning by optimizing the model to reconstruct the masked part using the visible context, which has been commonly used in natural language processing \cite{Devlin2018BERT,raffel2020t5} and computer vision \cite{He2022-MAE,xie2022simmim,li2023flip}. However, \citet{dong2023simmtm} found that randomly masking a portion of time points will seriously distort vital temporal correlations of time series, making the reconstruction task too difficult to guide representation learning. The other paradigm, contrastive learning, excels in instance-level representation learning, which optimizes the model to identify positive samples from negative ones \cite{tang2020exploring-simCLR}. A significant criticism of these contrastive approaches is their reliance on careful augmentations selection to learn useful invariances \cite{xiao2020should}, which is even harder in time series due to the scarcity of widely accepted and effective data augmentation methods \cite{wen2020time}. Also, the instance-level modeling design may fail in capturing fine-grained temporal variations, limiting their practicality to downstream tasks. 

We note a crucial distinction of time series from images or languages, as each time step consists of only a finite number of scalar values. This implies that the most vital information in time series is preserved in the temporal correlations, highlighting the importance of temporal modeling \cite{nie2022time-patch,wu2022timesnet}. Therefore, \emph{the critical point of time series pre-training is optimizing encoders to accurately capture temporal correlations}, which has not been adequately addressed in previous masking or contrastive methods.

To address the insufficiency in temporal modeling, we present TimeSiam, a simple yet effective self-supervised pre-training framework. Unlike prior, as shown in Figure \ref{fig:mask_schedule}, TimeSiam proposes to sample pairs of subseries across different timestamps from the same time series, termed ``\emph{Siamese subseries}''. Then, it leverages Siamese networks as encoders to capture correlations between temporally distanced subseries. With simple data augmentation such as masking, TimeSiam further improves the diversity and distinctiveness of Siamese subseries, which natively derives a past-to-current reconstruction task, thereby enforcing the encoder to learn temporally related information and capture correlations among past and current series. Besides, to cover different distanced Siamese subseries, we propose learnable lineage embeddings to enhance the encoder capacity for learning diverse time-dependent representations. Eventually, a decoder that integrates cross-attention and self-attention mechanisms is applied to ensure a precise reconstruction of the \lms{(masked) Siamese} subseries.

\lms{Importantly}, TimeSiam is not constrained by proximity information in the time series. Instead, benefiting from our Siamese subseries sampling procedure, it can effectively model the correlation among distanced subseries, which empowers the model with a more thorough understanding of the whole time series. With the above designs, TimeSiam \lms{remains simple but} achieves consistent state-of-the-art \lms{against prior} time series pre-training methods across various downstream tasks, including time series forecasting and classification, covering both in- and cross-domain settings. Overall, our contributions are summarized as follows:
\begin{itemize}
    \item In the spirit of learning temporal correlations, we propose TimeSiam, a simple but effective pre-training framework that leverages Siamese networks to capture correlations among temporally distanced subseries.
    \item With Siamese encoders to reconstruct current masked subseries based on past observation and lineage embeddings to capture subseries disparity, TimeSiam can learn diverse time-dependent representations.
    \item TimeSiam achieves consistent state-of-the-art fine-tuning performance across thirteen standard benchmarks, excelling in various time series analysis tasks.
\end{itemize}


\section{Related Work}
\subsection{Time Series Self-supervised Pre-training}

Self-supervised pre-training has demonstrated its ability to learn valuable and generalizable representations from large-scale unlabeled datasets in various domains, such as natural language processing (NLP) \cite{Devlin2018BERT,radford2019languageGPT-2,raffel2020t5, Brown2020GPT-1, Gao2021GPT-3} and computer vision (CV) \cite{he2020momentum, liu2021self, xie2022simmim, He2022-MAE}, which can significantly reduce labeling expenses and benefit diverse downstream tasks. Recently, self-supervised pre-training has empowered many breakthroughs in time series analysis by introducing well-established techniques into time series, such as masked modeling and contrastive learning.

\paragraph{Masked Modeling} As a fundamental technique in self-supervised pre-training methods, masked modeling enables deep models to learn essential representations by reconstructing masked parts from the visible context. Drawing inspiration from notable advances in NLP and CV, extensive time series pre-training approaches focus on time series masked modeling, which helps the model learn effective time series representations to facilitate various downstream analysis tasks. For instance, TST \citep{zerveas2021transformer-TST} and Ti-MAE \citep{li2023ti-TiMAE} propose to randomly mask segments and points in time series and pre-train the model with the reconstruction task. \camera{PatchTST \cite{nie2022time-patch} divides the temporal dimension into multiple patches and treats time series as independent variates. Additionally, it incorporates a non-overlapping patch-level masked self-supervised strategy for temporal representation learning}. HiMTM \citep{zhao2024himtm} proposes a novel hierarchical masked time series pre-training method to capture the multi-scale nature of time series. Additionally, SimMTM \citep{dong2023simmtm} introduces a multi-masking modeling paradigm, which reconstructs original time series through the weighted aggregation of multiple masked time series, thus being able to learn both points-wise and series-wise temporal representations. Despite these advances, all of these approaches solely focus on the modeling of one individual time series, disregarding the intrinsic temporal correlations and dynamical variations of the whole time series. In contrast, TimeSiam proposes to reconstruct the current sub-series based on past observations, which can naturally integrate the time-dependent information during reconstruction pre-training.

\paragraph{Contrastive Learning} Unlike masked modeling, this approach enables model pre-training by optimizing the similarity among instance-level representations. It leverages different data augmentations to construct positive and negative pairs from data, where positive pairs are optimized to be close to each other and negative pairs are encouraged to be distant from each other during pre-training \cite{tang2020exploring-simCLR,he2020momentum,chen2021exploring,gao2021simcse}. Current time series contrastive learning methods are mainly based on diverse data augmentations tailored to the domain-specific characteristics of time series. \camera{CPC \cite{oord2018cpc} introduced contrastive predictive coding, which uses model-predicted timesteps as positive samples and randomly-sampled timesteps as negative samples to obtain advantageous time series representations for downstream tasks. \citet{franceschi2019jean} combined a causal dilated convolutions-based encoder with a novel triplet loss that employs time-based negative sampling. TNC \cite{tonekaboni2021tnc} learns the representations by ensuring that signals from within a neighborhood are distinguishable from the distribution of non-neighboring signals in the latent space using a debiased contrastive loss.} TS2Vec \cite{Yue2022-TS2Vec} divides time series into patches, defining contrastive tasks at both the individual instance and patch levels. Mixing-up \cite{wickstrom2022mixing-Mixing-up} exploits a data augmentation scheme in which new samples are generated by mixing two data samples. LaST \cite{wang2022learning-LaST} aims to separate seasonal and trend components in time series data within the latent space. Additionally, CoST \cite{woo2022cost-CoST} utilizes contrastive losses in both time and frequency domains to learn distinct seasonal and trend representations. Furthermore, TF-C \cite{Zhang2022-TF-C} introduces a novel time-frequency consistency architecture, optimizing for proximity between time-based and frequency-based representations of the same data sample. However, existing contrastive learning methods for time series heavily rely on intricate data augmentation techniques to generate diverse views of the original data for self-supervision. Also, the instance-level representation learning may fall short in downstream low-level tasks. In TimeSiam, we utilize the native temporally distanced subseries to build reconstruction tasks, thereby freeing from complex augmentation techniques and also considering the low-level representation.

\subsection{Siamese Networks}

Siamese networks \cite{bromley1993signature} are particular neural network architectures with shared model parameters. This design makes Siamese networks well-suited for comparing and distinguishing two input samples \camera{based on a single neural network}. They have been widely used in contrastive learning to model the relationship between paired samples \cite{chen2021exploring}. The combination of Siamese networks and contrastive learning has been widely used in many applications, particularly in tasks requiring instance-level representations \cite{chen2020simple,he2020momentum, wang2023contrast}. However, in the field of time series pre-training, this combination generally focuses on recognizing subtle differences between various augmented views of the series itself, overlooking the essence of time series, that is temporal correlation modeling. In this paper, we explore using shared-weight Siamese autoencoders to establish correlations between past and current subseries. This methodology enables a more efficient understanding of temporal relations in time series and enforces the model to learn time-dependent representations.

\begin{figure*}[t]
\begin{center}
    \centerline{\includegraphics[width=\textwidth]{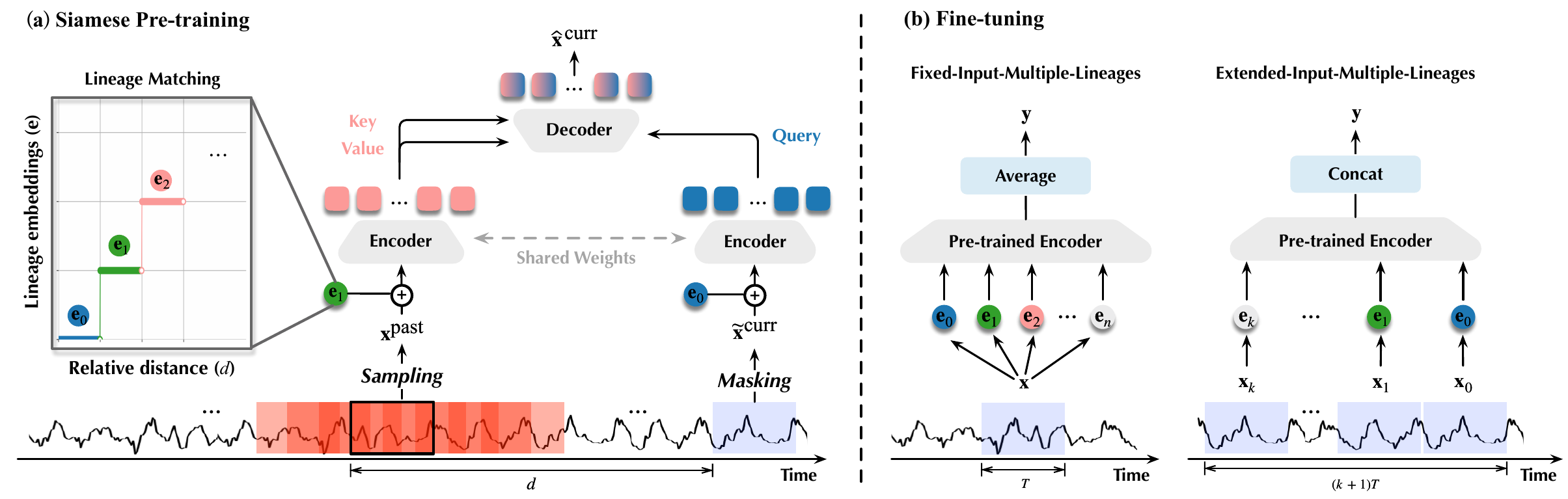}}
     \vspace{-5pt}
    \caption{The overall design of TimeSiam, which establishes correlations between subseries randomly sampled from different timestamps using Siamese encoders. It integrates learnable lineage embeddings to enhance the capacity for temporal-related representation learning.}
    \label{fig:overall}
\end{center}
\vspace{-25pt}
\end{figure*}

\vspace{-5pt}
\section{TimeSiam}
To enhance the time-dependent representation learning, TimeSiam is designed to capture correlations between temporally distant subseries based on Siamese networks. This framework can natively derive a past-to-current reconstruction task with simple masked augmentation. In addition, learnable lineage embeddings are incorporated to dynamically capture the disparity among different distanced subseries pairs, which can enhance the model's capacity to cover different temporal correlations. \lms{Hereafter}, we will detail the pre-training and fine-tuning stages in TimeSiam.

\subsection{Pre-training}
TimeSiam pre-training involves the following \update{two} modules: Siamese subseries sampling \update{and} Siamese modeling.

\vspace{-5pt}
\paragraph{Siamese Subseries Sampling}
Typically, previous time series pre-training approaches focus solely on modeling the individual series itself, neglecting the inherent correlations among temporally related time series. This deficiency in the pre-training phase will lead to insufficient extraction of generalizable time-dependent representations. In contrast, our TimeSiam is designed to focus on modeling temporal correlations of subseries across different timestamps, capturing the intrinsic time-correlated information of time series.

As shown in Figure \ref{fig:overall}, we construct Siamese subseries pairs by randomly sampling a past sample $\mathbf{x}^{\text{past}}$ \lms{preceding} the current sample $\mathbf{x}^{\text{curr}}$ in the same time series. Each sample in a Siamese pair, termed ``\emph{Siamese subseries}'' each other, contains $T$ timestamps and $C$ observed variables. Notably, one $\mathbf{x}^{\text{curr}}$ can correspond to multiple $\mathbf{x}^{\text{past}}$ subseries due to the random sampling process. We focus on constructing correlations and capturing temporal variations between these Siamese subseries, which benefits intrinsically time-dependent representation learning during pre-training. The relative distance between the past and current subseries, denoted as $d$, is crucial in representing the correlation and disparities between Siamese subseries. Furthermore, we adopt a simple masking augmentation to generate augmented current subseries $\widetilde{{\mathbf{x}}}^{\text{curr}}$ that further improves the diversity and the disparity of Siamese subseries pairs, ensuring a more robust and sufficient pre-training phase. The above process can be formalized as follows:
\begin{equation}
\begin{split}
(\mathbf{x}^{\text{past}}, \widetilde{\mathbf{x}}^{\text{curr}})&=\operatorname{Mask-Augment}\left((\mathbf{x}^{\text{past}}, \mathbf{x}^{\text{curr}})\right).
\end{split}
\end{equation}

\vspace{-5pt}
\paragraph{Siamese Modeling}

After constructing mask-augmented Siamese pairs, as shown in Figure \ref{fig:overall}, we further integrate learnable lineage embeddings during pre-training to effectively capture the disparity among different Siamese pairs. This design can enhance the model's capacity to extract diverse temporal-related representations. Given $N$ learnable lineage embeddings $\{\mathbf{e}^{\text{lineage}}_{i}\}_{i=1}^{N},\mathbf{e}^{\text{lineage}}_{i}\in\mathbb{R}^{1\times D}$ and \camera{$D$ represents the dimension of lineage embeddings}. For the past sample $\mathbf{{x}}^{\text{past}}$, we apply the $\operatorname{LineageMatching}(\cdot)$ function to dynamically match a certain lineage embedding based on its temporal distance $d$ to the current series. As for the current sample $\mathbf{\widetilde{x}}^{\text{curr}}$, we use a special lineage embedding to represent a degeneration situation as $d=0$:
\begin{equation}
\begin{split}
\mathbf{e}^{\text{lineage}}_{i} &= \operatorname{LineageMatching}(d) \\
\mathbf{z}^{\text{past}} &= \operatorname{Embed}(\mathbf{x}^{\text{past}}) \oplus \mathbf{e}^{\text{lineage}}_{i} \\
\mathbf{\widetilde{z}}^{\text{curr}} &= \operatorname{Embed}(\mathbf{\widetilde{x}}^{\text{curr}}) \oplus \mathbf{e}^{\text{lineage}}_{0}, 
\end{split}
\end{equation}
where $\mathbf{e}^{\text{lineage}}_{0} \in \mathbb{R}^{1\times D}$ is the specific embedding for current subseries and $\mathbf{z}^{\text{past}},\mathbf{\widetilde{z}}^{\text{curr}}\in\mathbb{R}^{T\times D}$ \lms{denote} the embedded Siamese features. \camera{Different base models correspond to different $\operatorname{Embed}(\cdot)$. Regarding PatchTST \cite{nie2022time-patch}, the patch-wise embedding function $\operatorname{PatchEmbed}(\cdot)$ is used to divide each variable into several patches and each patch is mapped to a patch token. As for iTransformer \cite{liu2023itransformer}, it uses the variable-wise embedding $\operatorname{VariateEmbed}(\cdot)$ and maps the entire variable into a temporal token.} Note that lineage embeddings are used to identify the temporal distance between Siamese subseries. It is shared along the time dimension when being added to Siamese subseries features. Here, $\oplus$ represents the addition operation with the temporal dimension broadcast.

Next, TimeSiam utilizes Siamese encoders to process pairs of Siamese pair features, which can be instantiated as advanced time series models, e.g.~PatchTST \cite{nie2022time-patch} or iTransformer \cite{liu2023itransformer}. After the Siamese encoder layer, we can obtain pairs of representations of past and masked current subseries \update{as follows:}
\begin{equation}
\begin{split}
\mathbf{h}^{\text{past}}_{e} = \operatorname{Encoder}(\mathbf{z}^{\text{past}}),\
\mathbf{\widetilde{h}}^{\text{curr}}_{e} = \operatorname{Encoder}(\mathbf{\widetilde{z}}^{\text{curr}}),
\end{split}
\end{equation}
where $\mathbf{h}^{\text{past}}_{e},\mathbf{\widetilde{h}}^{\text{curr}}_{e}\in\mathbb{R}^{T\times D}$ are from \lms{the} Siamese encoder.

Note that our Siamese sampling strategy natively derives a past-to-current reconstruction task. As shown in Figure~\ref{fig:overall}, we use a decoder that integrates cross-attention and self-attention mechanisms \cite{vaswani2017attention-Transformer} to incorporate past information into the current subseries for reconstruction, which can inherently capture the temporal correlations. Besides, this design can also enrich the limited context of masked current series to ensure accurate reconstruction for representation learning. Concretely, ${\mathbf{\widetilde{h}}}^{\text{curr}}_{e}$ serves as the query, and $\mathbf{h}^{\text{past}}_{e}$ acts as both the key and value, generating \update{the decoder representation of the current time subseries}, denotes as $\mathbf{\widehat{h}}_{d}$. This representation undergoes further refinement through a self-attention layer and a Feed-Forward Network (FFN). We formalize the decoder process as follows:
\begin{equation}
\begin{aligned}
\mathbf{\widehat{h}}_{d} &= \operatorname{LayerNorm}\Big({\mathbf{\widetilde{h}}}^{\text{curr}}_{e} + \operatorname{Cross-Attn}\big({\mathbf{\widetilde{h}}}^{\text{curr}}_{e}, \mathbf{h}^{\text{past}}_{e}, \mathbf{h}^{\text{past}}_{e}\big)\Big) \\
\mathbf{h}^{\prime}_{d} &= \operatorname{LayerNorm}\Big( \mathbf{\widehat{h}}_{d} + \operatorname{Self-Attn}\big(\mathbf{\widehat{h}}_{d}, \mathbf{\widehat{h}}_{d}, \mathbf{\widehat{h}}_{d} \Big) \\
\mathbf{h}_{d} &= \operatorname{LayerNorm}\Big( \mathbf{h}_{d}^{\prime} + \operatorname{FFN}\big(\mathbf{h}_{d}^{\prime}\big) \Big).
\end{aligned}
\end{equation}
\lms{We summarize this process as $\mathbf{h}_{d} = \operatorname{Decoder}({\mathbf{\widetilde{h}}}^{\text{curr}}_{e}, \mathbf{h}^{\text{past}}_{e}\big)$.}
Finally, the output of the decoder $\mathbf{h}_{d}\in\mathbb{R}^{T\times D}$ is used to reconstruct the masked current subseries through a linear projection layer, which can be formalized as:
\begin{equation}
\mathbf{\widehat{x}}^{\text{curr}}=\operatorname{Projector}(\mathbf{h}_{d}).
\end{equation}
Benefiting from our design, TimeSiam can be supervised by a simple reconstruction loss function and inherently learn time-dependent representations \lms{by} past-to-current temporal correlation modeling. The loss \lms{for each Siamese pair} is
\begin{equation}
     {\cal L}_{\text{reconstruction}}=\|\mathbf{x}^{\text{curr}}-\widehat{\mathbf{x}}^{\text{curr}}\|_{2}^{2}.
\end{equation}

\subsection{Fine-tuning}
Under the cooperation of lineage embeddings, the pre-trained Siamese encoder can capture diverse temporal-related representations under different lineage embeddings. As demonstrated in Figure \ref{fig:overall}(b), this advantage can further derive two types of fine-tuning paradigms, covering both fixed and extended input series settings.

\paragraph{Fixed-Input-Multiple-Lineages}

In a standard fine-tuning scenario, a sample typically generates only one type of representation, which seriously limits the capacity of the pre-trained encoder. In contrast, TimeSiam innovatively pre-trains Siamese encoders with diverse lineage embeddings to capture different distanced temporal correlations, which allows TimeSiam to derive diverse representations with different lineages for the same input series. This procedure subtly releases the capacity of the pre-trained model and enhances the diversity of extracted representations. Given \lms{an} input series $\mathbf{x}\in\mathbb{R}^{T\times C}$, this process can be written as
\begin{equation}
\begin{split}
\overline{\mathbf{h}}_{e} &= \operatorname{Average}\left(\mathbf{h}_{e, 0},\mathbf{h}_{e, 1}, ... \mathbf{h}_{e, n} \right), \\
\text{where} \ \mathbf{h}_{e, i} &= \operatorname{Encoder}\left( \operatorname{Embed}(\mathbf{x}) \oplus \mathbf{e}^{\text{lineage}}_{i} \right).
\end{split}
\end{equation}
The final output $\overline{\mathbf{h}}_{e}\in\mathbb{R}^{T\times D}$ is an ensemble of a \lms{set} of temporal representations derived from \update{the same input series} $\mathbf{x}$ but with different lineage embeddings $\mathbf{e}^{\text{lineage}}_{i}$, which can cover diverse temporal-related information of input series.

\vspace{-5pt}
\paragraph{Extended-Input-Multiple-Lineages} Note that in the fine-tuning stage, the model may receive longer records than the pre-training series. Given a $(k+1)T$-length input $(\mathbf{x}_{k},\cdots,\mathbf{x}_{1},\mathbf{x}_{0}), \mathbf{x}_{i}\in\mathbb{R}^{T\times C}$, previous time series pre-training methods have to adopt the same encoder to different segments, which clearly overlooks the chronological order of extended series. \lms{Desirably}, in TimeSiam, we can leverage multiple lineage embeddings trained under different temporal distanced pairs to different segments, which can natively conserve the \lms{temporal} order of different segments. This advantage is achieved by associating each segment with its respective lineage embedding:
\begin{equation}
\begin{split}\label{equ:extended}
\overline{\mathbf{h}}_{e} &= \operatorname{Concat}\left( \mathbf{h}_{e, 0},\mathbf{h}_{e, 1}, \ldots, \mathbf{h}_{e, k} \right), \\
\text{where} \ \mathbf{h}_{e, i} &= \operatorname{Encoder}\left( \operatorname{Embed}(\mathbf{x}_{i}) \oplus \mathbf{e}^{\text{lineage}}_{{\operatorname{\text{\tiny{LineageMatching}}}(\lms{\tiny{iT}})}} \right).
\end{split}
\end{equation}
Here $\overline{\mathbf{h}}_{e}\in\mathbb{R}^{(k+1)T\times D}$ \lms{denotes} the extracted representation for extended input series.

To align the experiment setting with previous work \cite{dong2023simmtm}, our experiments are based on the Fixed-Input-Multiple-Lineages setting except for special clarification.

\section{Experiments}

We perform extensive experiments across two mainstream time series analysis tasks: forecasting and classification, covering both in- and cross-domain settings.

\subsection{Experimental Setup}

\paragraph{Datasets}

We summarize the experimental benchmarks in Table \ref{tab:benchmarks}, encompassing eleven well-established datasets and two newly constructed datasets, which cover two primary tasks in time series analysis: forecasting and classification. It is worth noting that to further demonstrate the pre-training benefits under large and diverse data, we employ the TSLD dataset, which is constructed by merging time series datasets from multiple domains that are nonoverlapping with the other datasets. This allows us to explore cross-domain transfer scenarios with large-scale pre-training data. Please refer to Appendix \ref{app:datasets} for a more comprehensive description.

\begin{table}[h]
        \vspace{-5pt}
	\caption{{Summary of experiment benchmarks, where TSLD-500M and TSLD-1G are newly constructed from diverse domains.}}
	\label{tab:benchmarks}
	\vskip 0.05in
	\centering
	\begin{small}
        \renewcommand{\multirowsetup}{\centering}
        \setlength{\tabcolsep}{4.0pt}
        \renewcommand\arraystretch{1.2}
        \scalebox{1}{
        \begin{tabular}{clcc}
            \toprule
            \scalebox{1}{\textsc{Tasks}} & \scalebox{1}{\textsc{Datasets}} & \scalebox{1}{\textsc{Domain}} & \scalebox{1}{\textsc{Examples}} \\
            \midrule
            \multirow{7}{*}{\scalebox{1}{Forecasting}} & \scalebox{1}{ETT (4 subsets)} & \scalebox{1}{Electricity} & 14.3K \\
                        & \scalebox{1}{Weather} & \scalebox{1}{Weather} & 52.7K \\
                        & \scalebox{1}{Electricity} & \scalebox{1}{Electricity} & 26.3K \\
                        & \scalebox{1}{Traffic} & \scalebox{1}{Transportation} & 17.5K \\      
                        & \scalebox{1}{Exchange} & \scalebox{1}{Finance} & 7.6K \\
                        \cmidrule(lr){2-4}
                        & \scalebox{1}{TSLD-500M} & \scalebox{1}{Multiple} & 412.6K \\
                        & \scalebox{1}{TSLD-1G} & \scalebox{1}{Multiple} & 13.9M \\
            \midrule
            \multirow{3}{*}{\scalebox{1}{Classification}} & \scalebox{1}{AD} & \scalebox{1.0}{EEG} & 5.97K\\
                & \scalebox{1}{TDBrain} & \scalebox{1}{EEG} & 11.9K\\
                & \scalebox{1}{PTB} & \scalebox{1}{ECG} & 62.4K \\
            \bottomrule
        \end{tabular}}
	\end{small}
\end{table}

\begin{table*}[!th]
        \vspace{-5pt}
	\caption{\update{In-domain fine-tuning for time series forecasting. Siamese encoders are both pre-trained and fine-tuned on the same dataset. Results are the average Mean Squared Error (MSE), calculated from forecasts made for four future lengths $O \in \{96,192,336,720\}$, based on the past 96 time points. A smaller MSE indicates a better prediction. Full results are presented in Appendix \ref{app:full results}.}}
 	\label{tab:forecasting_indomain_shot}
        \vspace{-15pt}
	\vskip 0.3in
        \setlength{\tabcolsep}{5.0pt}
        \renewcommand\arraystretch{1.2}
	\centering
	\begin{small}
        \begin{sc}
        \renewcommand{\multirowsetup}{\centering}
        \begin{tabular}{c|l|cccccccc}
        \toprule
        Encoder & Method & ETTh1 & ETTh2 & ETTm1 & ETTm2 & Weather & Exchange & ECL & Traffic \\
        \midrule
        \multirow{12}{*}{PatchTST} & \textcolor{gray}{Random init.} & \textcolor{gray}{0.473} & \textcolor{gray}{0.385} & \textcolor{gray}{0.390} & \textcolor{gray}{0.285} & \textcolor{gray}{0.259} & \textcolor{gray}{0.367} & \textcolor{gray}{0.216} & \textcolor{gray}{0.490}\\
        \cmidrule(lr){2-10}
        & CPC (\citeyear{oord2018cpc}) & 0.440 & 0.401 & 0.389 & 0.290 & 0.272 & 0.368 & 0.220 & 0.504 \\
        & TNC (\citeyear{tonekaboni2021tnc}) & 0.445 & 0.379 & 0.386 & 0.287 & 0.270 & 0.362 & 0.212 & 0.501 \\
        & TS2Vec (\citeyear{Yue2022-TS2Vec}) & 0.456 & 0.376 & 0.393 & 0.289 & 0.256 & 0.363 & 0.199 & 0.472 \\
        & CoST (\citeyear{woo2022cost-CoST}) & 0.457 & 0.374 & 0.395 & 0.286 & 0.253 & 0.364 & 0.203 & 0.480 \\
        & LaST (\citeyear{wang2022learning-LaST}) & 0.479 & 0.385 & 0.398 & 0.285 & 0.252 & 0.433 & 0.207 & 0.520 \\
        & TF-C (\citeyear{Zhang2022-TF-C}) & 0.453 & 0.378 & 0.389 & 0.281 & 0.257 & 0.362 & 0.202 & 0.487 \\
        & TST (\citeyear{zerveas2021transformer-TST}) & 0.452 & 0.383 & 0.380 & 0.288 & 0.259 & 0.385 & 0.197 & 0.486 \\
        & Ti-MAE (\citeyear{li2023ti-TiMAE}) & 0.448 & 0.379 & 0.384 & 0.279 & 0.257 & 0.370 & 0.196 & 0.481 \\
        & PatchTST$^{\dag}$ (\citeyear{nie2022time-patch}) & 0.442 & 0.381 & 0.379 & 0.285 & 0.267 & 0.358 & 0.200 & 0.484 \\
        & SimMTM (\citeyear{dong2023simmtm}) & 0.440 & 0.382 & 0.377 & 0.285 & 0.256 & 0.361 & 0.192 & 0.466 \\
        \cmidrule(lr){2-10}
        & \textbf{TimeSiam} & \textbf{0.429} & \textbf{0.373} & \textbf{0.374} & \textbf{0.279} & \textbf{0.252} & \textbf{0.353} & \textbf{0.189} & \textbf{0.453} \\
        \midrule
        \multirow{10}{*}{iTransformer} & \textcolor{gray}{Random init.} & \textcolor{gray}{0.454} & \textcolor{gray}{0.383} & \textcolor{gray}{0.407} & \textcolor{gray}{0.288} & \textcolor{gray}{0.258}& \textcolor{gray}{0.365} & \textcolor{gray}{0.178} & \textcolor{gray}{0.428} \\
        \cmidrule(lr){2-10}
        & TS2Vec (\citeyear{Yue2022-TS2Vec}) & 0.474 & 0.379 & 0.411 & 0.290 & 0.264 & 0.364 & 0.246 & 0.485 \\
        & CoST (\citeyear{woo2022cost-CoST}) & 0.472 & 0.386 & 0.411 & 0.294 & 0.269 & 0.366 & 0.252 & 0.529\\
        & LaST (\citeyear{wang2022learning-LaST}) & 0.465 & 0.386 & 0.400 & 0.302 & 0.262 & 0.386 & 0.237 & 0.477\\
        & TF-C (\citeyear{Zhang2022-TF-C}) & 0.450 & 0.379 & 0.403 & 0.292 & 0.265 & 0.372 & 0.222 & 0.432 \\
        & TST (\citeyear{zerveas2021transformer-TST}) & 0.447 & 0.376 & 0.399 & 0.291 & 0.261 & 0.363 & 0.228 & 0.438 \\
        & Ti-MAE (\citeyear{li2023ti-TiMAE}) & 0.448 & 0.378 & 0.399 & 0.289 & 0.257 & 0.366 & 0.217 & 0.430 \\
        & SimMTM (\citeyear{dong2023simmtm}) & 0.445 & 0.376 & 0.397 & 0.286 & 0.259 & 0.358 & 0.179 & 0.426 \\
        \cmidrule(lr){2-10}
        & \textbf{TimeSiam} & \textbf{0.440} & \textbf{0.371} & \textbf{0.390} & \textbf{0.284} & \textbf{0.256} & \textbf{0.355} & \textbf{0.175} & \textbf{0.420} \\
        \bottomrule
        \end{tabular}
        \end{sc}
	\end{small}
    \vspace{-5pt}
\end{table*}

\begin{table*}[th]
	\caption{\update{Cross-domain fine-tuning for time series forecasting. Siamese encoders are pre-trained on the TSLD-1G dataset and fine-tuned on various target datasets. Results are the average Mean Squared Error (MSE) for four future lengths $O \in \{96,192,336,720\}$, based on the past 96 time points. A smaller MSE indicates a better prediction. Full results are detailed in Appendix \ref{app:full results}.}}
  	\label{tab:forecasting_crossdomain_shot}
        \vspace{-15pt}
	\vskip 0.3in
        \setlength{\tabcolsep}{6.0pt}
        \renewcommand\arraystretch{1.2}
	\centering
	\begin{small}
        \begin{sc}
        \renewcommand{\multirowsetup}{\centering}
        \begin{tabular}{l|l|cccccccc}
        \toprule
        Encoder & Method & ETTh1 & ETTh2 & ETTm1 & ETTm2 & Weather & Exchange & ECL & Traffic \\
        \midrule
        \multirow{7}{*}{PatchTST} & \textcolor{gray}{Random init.} & \textcolor{gray}{0.473} & \textcolor{gray}{0.385} & \textcolor{gray}{0.390} & \textcolor{gray}{0.285} & \textcolor{gray}{0.259} & \textcolor{gray}{0.367} & \textcolor{gray}{0.216}  & \textcolor{gray}{0.490} \\
        \cmidrule(lr){2-10}
        & TS2Vec (\citeyear{Yue2022-TS2Vec}) & 0.441 & 0.375 & 0.380 & 0.291 & 0.256 & 0.365 & 0.217 & 0.528 \\
        & TF-C (\citeyear{Zhang2022-TF-C}) & 0.437 & 0.378 & 0.386 & \textbf{0.282} & 0.264 & 0.360 & 0.218 & 0.543 \\
        & TST (\citeyear{zerveas2021transformer-TST}) & 0.434 & 0.384 & 0.387 & 0.303 & 0.263 & 0.365 & 0.220 & 0.514 \\
        & Ti-MAE (\citeyear{li2023ti-TiMAE}) & 0.435 & 0.374 & 0.380 & 0.294 & 0.256 & 0.362 & 0.218 & 0.515\\
        & SimMTM (\citeyear{dong2023simmtm}) & 0.429 & 0.380 & 0.375 & 0.287 & 0.252 & 0.365 & 0.213 & 0.459 \\
        \cmidrule(lr){2-10}
        & \textbf{TimeSiam} & \textbf{0.425} & \textbf{0.374} & \textbf{0.371} & 0.286 & \textbf{0.251}  & \textbf{0.360} & \textbf{0.188} & \textbf{0.454}\\
        \bottomrule
        \end{tabular}
        \end{sc}
	\end{small}
     \vspace{-5pt}
\end{table*}

\vspace{-5pt}
\paragraph{Backbone}
We use the advanced time series models across various tasks as \update{Siamese encoders} to evaluate the efficacy of our pre-training methods. In particular, we utilized iTransformer \cite{liu2023itransformer} and PatchTST \cite{nie2022time-patch} as the encoder for time series forecasting \camera{following their original configurations. The patch length and stride were both set to 12 without any overlap}. TCN \cite{bai2018tcn} is used as the backbone for the classification task \cite{wang2023contrast}. \update{Note that to ensure a fair comparison, we unify the encoder backbone of our model and all the baselines. The results with a unified encoder generally surpass the results reported by themselves across all baselines.}

\vspace{-5pt}
\paragraph{Baselines}
We compare our TimeSiam with eight \update{advanced} self-supervised time series pre-training baselines \update{under} the in-domain setting, including contrastive learning methods: COMET \citeyearpar{wang2023contrast}, TF-C \citeyearpar{Zhang2022-TF-C}, LaST \citeyearpar{wang2022learning-LaST}, CoST \citeyearpar{woo2022cost-CoST}, TS2Vec \citeyearpar{Yue2022-TS2Vec}, \camera{TNC \citeyearpar{tonekaboni2021tnc}, CPC \citeyearpar{oord2018cpc}} and \update{masked modeling methods}: SimMTM~\citeyearpar{dong2023simmtm}, Ti-MAE \citeyearpar{li2023ti-TiMAE}, TST \citeyearpar{zerveas2021transformer-TST}, \camera{and a patch-wise self-surpervised masked modeling method, PatchTST, proposed by \cite{nie2022time-patch}}. 

\camera{It should be noted that some baselines such as CPC \cite{oord2018cpc}, TNC \cite{tonekaboni2021tnc}, etc. are not validated in the iTransformer backbone because their specific design based on internal modeling of series is not applicable to iTransformer \cite{liu2023itransformer}, which models the entire sequence as a whole temporal token and focuses on modeling relationships between different variables.} In the cross-domain setting, due to the large-scale dataset TSLD will bring huge experiment costs, we only select part of the baselines (the efficient ones) for comparisons. \update{Besides, COMET \citeyearpar{wang2023contrast} is specifically designed for medical time series and we also exclude it from cross-domain evaluation.} \camera{More implementation details can be found in Appendix \ref{app:implementation}.}

\subsection{Main Results}
\begin{figure}[h]
\begin{center}
    \centerline{\includegraphics[width=0.49\textwidth]{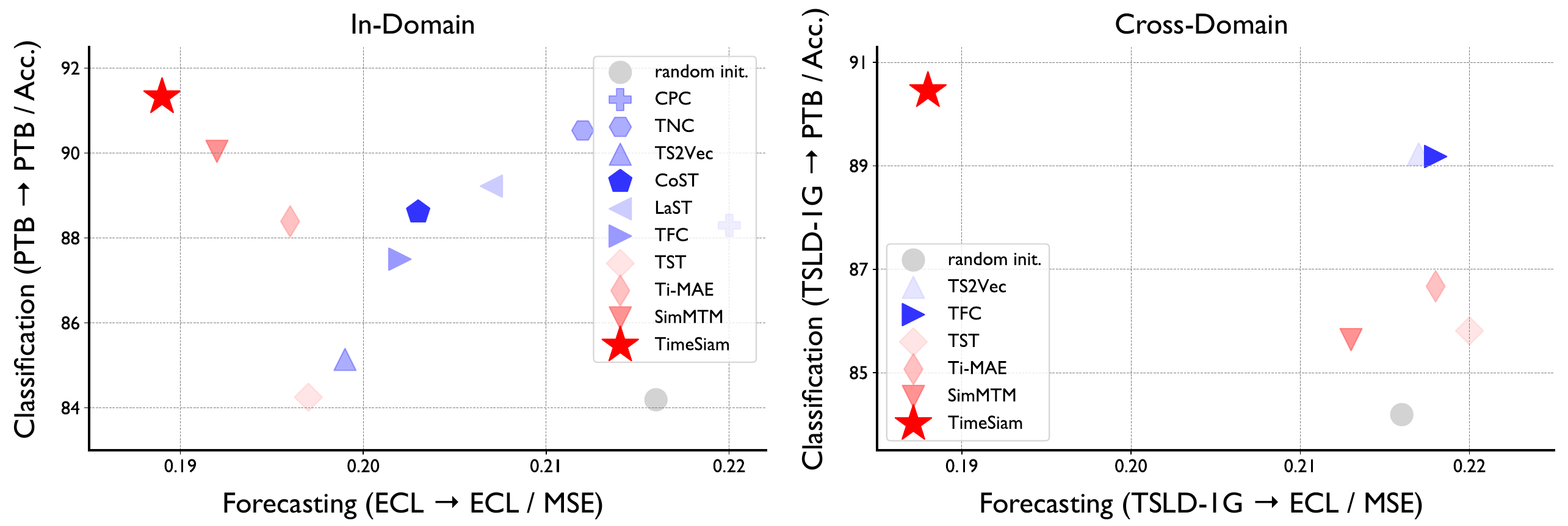}}
    \vspace{-5pt}
	\caption{Comparison of time series pre-training baselines for forecasting (MSE $\downarrow$) and classification (Accuracy $\uparrow$) tasks. This comparison included both contrastive-based and masking-based methods, covering both in- (left) and cross-domain (right) settings.}
	\label{fig:mainresult}
    \vspace{-30pt}
\end{center}
\end{figure}

As shown in Figure \ref{fig:mainresult},\camera{we summarize the performance of our TimeSiam \update{in} both in- and cross-domain scenarios for two mainstream time series analysis tasks: time series forecasting  ($x$-axis) and classification ($y$-axis)}. For each scenario, TimeSiam exhibits significant improvement over other established strong self-supervised baselines for time series.

\subsection{Forecasting}

\paragraph{In-domain}

We investigate the effectiveness of TimeSiam by integrating it with state-of-the-art time series forecasting models: PatchTST \cite{nie2022time-patch} and iTransformer \cite{liu2023itransformer}. As shown in Table \ref{tab:forecasting_indomain_shot}, TimeSiam can further enhance model performance, achieving an average MSE reduction of 5.7\% and 2.5\% across all forecasting benchmarks, even though these advanced models already exhibit excellent forecasting capabilities from random initialization. \camera{Significantly, it is evident that the masked modeling-based approach overall outperforms the contrastive-based approach in the forecasting task. This can be attributed to the advantages gained through series reconstruction to learn low-level temporal representation.} However, our TimeSiam still \update{performs the best forecasting capability} among all existing state-of-the-art self-supervised baseline methods.

\vspace{-5pt}

\paragraph{Cross-domain}
As shown in Table \ref{tab:forecasting_crossdomain_shot}, we use the TSLD-1G dataset, which contains larger scale time series samples from diverse domains, to validate the effectiveness of TimeSiam in a cross-domain transfer setting. \update{Note that this setting not only requires pre-training learning from large-scale data but also poses thorny challenges in handling mismatched data distribution.} The results \camera{consistently demonstrate that our TimeSiam significantly enhances the performance over training from random initialization covering all forecasting benchmarks,} achieves comparable results in the in-domain setting, and consistently outperforms other baseline methods. It is worth noting that the transfer results even show superior performance compared to the in-domain scenario in some datasets, such as TSLD-1G $\to$ \{ETTh1, ETTm1\}. This confirms the essential significance of using more large-scale and varied data for time series pre-training.

\begin{table}[t]
	\caption{{In-domain fine-tuning for time series classification}. The model is pre-trained on two EEG datasets: AD and TDBrain, and an ECG dataset: PTB, and then fine-tuned on the same dataset. Accuracy (\%) is recorded. See Appendix \ref{app:full results} for more details.}
	\label{tab:classification_indomain_short}
        \setlength{\tabcolsep}{10.0pt}
        \renewcommand\arraystretch{1.2}
	\vskip 0.1in
	\centering
	\begin{small}
		\begin{sc}
			\renewcommand{\multirowsetup}{\centering}
			\begin{tabular}{l|ccc}
				\toprule
                    Method & AD & TDBrain & PTB \\
				\midrule
				\textcolor{gray}{Random init.}& \textcolor{gray}{80.62} & \textcolor{gray}{79.08} & \textcolor{gray}{84.19} \\
                    \midrule
                    CPC (\citeyear{oord2018cpc}) & 77.40 & 85.19 & 88.30 \\
                    TNC (\citeyear{tonekaboni2021tnc}) & 78.58 & 85.21 & 90.53 \\
				TS2vec (\citeyear{Yue2022-TS2Vec}) & 81.26 & 80.21 & 85.14 \\
				CoST (\citeyear{woo2022cost-CoST}) & 73.87 & 83.86 & 88.61 \\
				LaST (\citeyear{wang2022learning-LaST}) & 72.63 & 85.13 & 89.22 \\
				TF-C (\citeyear{Zhang2022-TF-C}) & 75.31 & 66.62 & 87.50 \\
    			COMET (\citeyear{wang2023contrast}) & 84.50 & 85.47 & 87.84 \\
                    TST (\citeyear{zerveas2021transformer-TST}) & 81.50 & 83.22 & 84.25 \\
                    Ti-MAE (\citeyear{li2023ti-TiMAE}) & 80.70 & 88.16 & 88.39 \\
                    SimMTM (\citeyear{dong2023simmtm}) & 86.19 & 84.81 & 90.04 \\
				\midrule
    			\textbf{TimeSiam} &  \textbf{89.93} & \textbf{90.67} & \textbf{91.32} \\
				\bottomrule
			\end{tabular}
		\end{sc}
	\end{small}
       \vspace{-5pt}
\end{table}

\begin{table}[!h]
	\caption{Cross-domain fine-tuning for time series classification. The model is pre-trained on TSLD-1G dataset and fine-tuned on EEG dataset AD, TDBrain and ECG dataset PTB. Accuracy (\%) is recorded and further details can be found in Appendix \ref{app:full results}.}
	\label{tab:classification_crossdomain_short}
        \setlength{\tabcolsep}{10.5pt}
        \renewcommand\arraystretch{1.2}
	\vskip 0.1in
	\centering
	\begin{small}
		\begin{sc}
			\renewcommand{\multirowsetup}{\centering}
			\begin{tabular}{l|ccc}
				\toprule
                    Method & AD & TDBrain & PTB \\
				\midrule
				\textcolor{gray}{Random init.} & \textcolor{gray}{80.62} & \textcolor{gray}{79.08} & \textcolor{gray}{84.19} \\
                    \midrule
				TS2vec (\citeyear{Yue2022-TS2Vec}) & 80.59 & 85.58 & 89.23 \\
				TF-C (\citeyear{Zhang2022-TF-C}) & 87.98 & 82.84 & 89.18 \\
                    TST (\citeyear{zerveas2021transformer-TST}) & 82.60 & 83.65 & 85.81\\
                    Ti-MAE (\citeyear{li2023ti-TiMAE}) & 80.40 & 85.22 & 86.67 \\
                    SimMTM (\citeyear{dong2023simmtm}) & 87.74 & 85.29 & 85.64\\
				\midrule
    			\textbf{TimeSiam} & \textbf{90.47} & \textbf{86.26} & \textbf{90.45} \\
				\bottomrule
			\end{tabular}
		\end{sc}
          \vspace{-10pt}
	\end{small}
\end{table}

\begin{table*}[th]
     \setlength{\abovecaptionskip}{0.cm}
    \setlength{\belowcaptionskip}{-0.cm}
  \caption{Ablations on the Traffic benchmark, which are conducted under the in-domain forecasting setting. The length of the input series was fixed at 96 time points. The default setting is indicated by a grey bold marking. MSE averaged from 4 forecasting horizons $\{96,192,336,720\}$ is reported. The (c) notation in masking rules refers to channel-wise masking. More results can be found in Table \ref{tab:forecast_in_domain_abs_full}.}
  \vspace{5pt}
  \label{tab:main_ablation}
  \centering
  \begin{small}
  \renewcommand{\multirowsetup}{\centering}
  \setlength{\tabcolsep}{18pt}
  \renewcommand\arraystretch{1.2}
  \begin{tabular}{llcc|cccc}
    \toprule
    \multicolumn{4}{c}{(a) \textsc{Siamese Sampling}} & \multicolumn{4}{c}{(b) \textsc{Subseries Reconstruction}}  \\
    \multicolumn{2}{c}{\scalebox{1}{Sampling range $(\max d)$}} & \multicolumn{2}{c}{Lineage types $N$} & \multicolumn{2}{c}{Masking rules} & \multicolumn{2}{c}{Masked ratio} \\
    \cmidrule(lr){1-2}\cmidrule(lr){3-4}\cmidrule(lr){5-6}\cmidrule(lr){7-8}
     $w/o$ & 0.462 & $w/o$ & 0.457  & \text{binomial} & 0.459 & $w/o$ & 0.460 \\
     1 & 0.454 & 2 & 0.456 & \text{continuous} & 0.459 & 15\% & 0.459  \\
     3 & 0.454 & \textbf{3} & \cellcolor{gray!20}{\textbf{0.453}} & \text{mask last} & 0.461 & \textbf{25\%} & \cellcolor{gray!20}{\textbf{0.453}} \\
     \textbf{6} & \cellcolor{gray!20}{\textbf{0.453}} & 6 & 0.455 & \text{binomial (c)} & 0.457 & 50\% & 0.455 \\
     12 & 0.455 & & & \textbf{continuous (c)} & \cellcolor{gray!20}{\textbf{0.453}} & 75\% & 0.457 \\
    \bottomrule
  \end{tabular}
  \end{small}
  \vspace{-10pt}
\end{table*}

\subsection{Classification}

\paragraph{In-domain}

To further investigate the generalizability of the representations learned by TimeSiam, we examined the impact of in-domain pre-training on classification tasks within the medical domain, \update{following the setup in \cite{wang2023contrast}}. Results in Table \ref{tab:classification_indomain_short} demonstrate competitive outcomes achieved by both COMET \cite{wang2023contrast} and SimMTM \cite{dong2023simmtm}. This can be attributed to the elaborative designs of their approaches, where COMET incorporates domain-specific knowledge into its design and SimMTM models both high-level and low-level temporal representations. Compared with these competitive baselines, our TimeSiam, characterized by its simplicity and generality, consistently achieves remarkable results. \camera{In all classification benchmarks, our proposed TimeSiam consistently enhances the average classification accuracy by 11.5\% compared to random initialization, surpassing other baseline methods.}

\paragraph{Cross-domain}

\update{Furtherly}, we investigated the impact of using larger and more diverse pre-training datasets on time series classification tasks: TSLD-1G $\to$ \{AD, TDBrain, PTB\}. Generally, as shown in Table \ref{tab:classification_crossdomain_short}, pre-training will bring consistent promotion w.r.t. random initialization. \update{However, it is also observed that employing larger datasets from more diverse domains does not definitively show an advantage over the in-domain pre-trained models.} This result may come from the inherent differences between the domains of pre-training and fine-tuning under the cross-domain setting. Notably, even in this tough cross-domain setting, TimeSiam still surpasses the other baselines, demonstrating its capability to handle shifted data distributions.

\subsection{Ablation Studies}

We conduct extensive ablation studies to evaluate the effectiveness of various designs in TimeSiam, including \update{Siamese sampling and subseries reconstruction}.

\vspace{-10pt}
\paragraph{Siamese Sampling}

We explore the key hyper-parameters of Siamese modeling: the maximum sampling distance between the past and current Siamese subseries (Sampling range $\max d$) and the size of the lineage embedding set (Lineage types $N$). \camera{The maximum sampling range is determined by the length of the subseries ($T$) and a hyperparameter ($r$), resulting in $\max d = T \times r$}. Table \ref{tab:main_ablation}(a) demonstrates that integrating past-to-current Siamese modeling outperforms self-reconstruction modeling in time series pre-training. Furthermore, expanding the sampling range of Siamese subseries reasonably significantly enhances performance, underscoring the critical role of Siamese modeling in achieving optimal fine-tuning results.

\vspace{-10pt}
\paragraph{Subseries Reconstruction}

For subseries reconstruction, results in Table \ref{tab:main_ablation}(b) indicate that channel-wise masking significantly benefits Siamese modeling, especially when comparing the \emph{continuous} to channel-wise \emph{continuous (c)}. Both random and continuous masking work well, achieving lower mean squared error in the process. As for the masking ratio, we find that a high masking ratio of 75\% will significantly distort temporal variations, whereas a low mask ratio of 15\% overly simplifies the task, hindering effective temporal representation learning. Therefore, we adopt a default masking ratio of 25\%, same as~\cite{Devlin2018BERT}.

\paragraph{Lineage embeddings}

\camera{As shown in Table \ref{tab:embeddings}, we can observe that lineage embeddings play a vital role in enhancing forecasting performance. This comes from the inherent ability of lineage embeddings to distinguish temporal distances between subseries during the pre-training phase. As a result, they facilitate the learning of diverse temporal correlations. Consequently, the inclusion of lineage embeddings has been shown to be effective in improving performance when fine-tuning models for a variety of downstream tasks.}

\begin{figure*}[th]
\begin{center}
    \centerline{\includegraphics[width=1\textwidth]{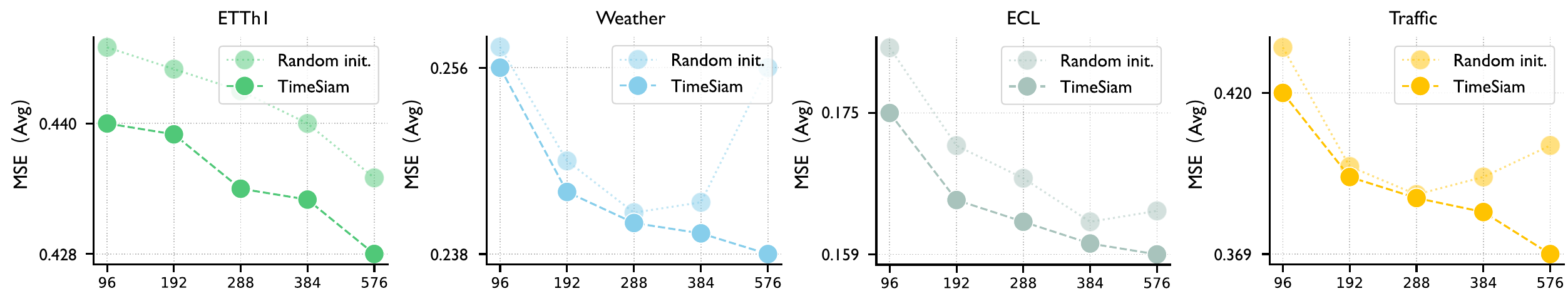}}
    \vspace{-5pt}
    \caption{Fine-tuning the pre-trained model to the inputs with extended length \{96, 192, 288, 384, 576\} based on iTransformer \cite{liu2023itransformer}. The MSE averaged from all predicted horizons $\{96,192,336,720\}$ is reported. Additional results are in the Appendix \ref{app:adapt_length_full_result}.}
    \vspace{-5pt}
    \label{fig:long_lookback}
    \vspace{-20pt}
\end{center}
\end{figure*}

\begin{table}[t]
  \vspace{-5pt}
    \setlength{\abovecaptionskip}{0.cm}
    \setlength{\belowcaptionskip}{-0.cm}
  \caption{Fine-tuning performance of TimeSiam under different pre-training data and model sizes. Relative improvement over random initialization (\%) is marked in green. See Appendix \ref{appdix:backbone} for details.}
  \vspace{5pt}
  \label{tab:large_data}
  \centering
  \begin{small}
  \renewcommand{\multirowsetup}{\centering}
  \setlength{\tabcolsep}{8pt}
  \renewcommand\arraystretch{1.2}
  \begin{tabular}{l|ll}
        \toprule
        \textsc{Pre-train} & \textsc{Traffic} & \textsc{ECL} \\
        \midrule
\textcolor{gray}{Random initiation} & \textcolor{gray}{0.490} & \textcolor{gray}{0.216} \\
TimeSiam $\text{in-domain}$ & 0.453 & 0.189 \\
        \midrule
        TimeSiam$_{\text{Base}}$ TSLD$_{\text{500M}}$ & 0.462 \textcolor{mygreen}{(+5.7)} & 0.189 \textcolor{mygreen}{(+12.5)} \\
        TimeSiam$_{\text{Base}}$ TSLD$_{\text{1G}}$ & 0.454 \textcolor{mygreen}{(+7.4)} & 0.188 \textcolor{mygreen}{(+12.5)} \\
        TimeSiam$_{\text{Large}}$ TSLD$_{\text{1G}}$ & 0.433 \textcolor{mygreen}{(+11.6)} & 0.185 \textcolor{mygreen}{(+14.4)} \\
        \bottomrule
  \end{tabular}
  \end{small}
  \vspace{-5pt}
\end{table}

\begin{table}[h]
  \vspace{-10pt}
     \setlength{\abovecaptionskip}{0.cm}
    \setlength{\belowcaptionskip}{-0.cm}
  \caption{Ablations on lineage embeddings. The MSE averaged from 4 forecasting horizons \{96, 192, 336, 720\} is reported here.}
  \vspace{5pt}
  \label{tab:embeddings}
  \centering
  \begin{small}
  \renewcommand{\multirowsetup}{\centering}
  \setlength{\tabcolsep}{5pt}
  \renewcommand\arraystretch{1.2}
  \begin{tabular}{l|ccc}
        \toprule
        \textsc{Datasets} & \textsc{Random init.} & \textsc{TimeSiam} & \textsc{w/o Lineage}\\
        \midrule
        ETTh1 & 0.473 & 0.429 & \textcolor{gray}{0.433 $\downarrow$} \\
        ETTm1 & 0.390 & 0.374 & \textcolor{gray}{0.378 $\downarrow$} \\
        Weather & 0.259 & 0.252 & \textcolor{gray}{0.256 $\downarrow$} \\
        Traffic & 0.490 & 0.453 & \textcolor{gray}{0.457 $\downarrow$} \\
        Exchange & 0.36 & 0.353 & \textcolor{gray}{0.365 $\downarrow$} \\
        \bottomrule
  \end{tabular}
  \end{small}
  \vspace{-5pt}
\end{table}

\subsection{Analysis Experiment}

\paragraph{Data Scale and Model Capacity}

One of the bottlenecks that block the development of time series pre-training is the lack of large-scale and diverse data for pre-training \cite{zhou2023one}. To investigate the influence of data scale on TimeSiam, we employed TimeSiam for pre-training on a larger dataset TSLD-\{0.5G, 1G\} along with different model sizes, followed by applying it to downstream time-series prediction tasks to assess fine-tuning effects. Results, illustrated in Table \ref{tab:large_data}, reveal that the model performance is promoted significantly, empowering by TimeSiam pre-training (TimeSiam vs. random init.). Furthermore, although $\text{TSLD}_{\text{500M}}$ does not exhibit a significant advantage over TimeSiam under the in-domain setting initially, we observed a marked enhancement in performance as the dataset size increased ($\text{TSLD}_{\text{1G}}$ vs. $\text{TSLD}_{\text{500M}}$), and $\text{TimeSiam}_{\text{Large}}$ significantly outperforms $\text{TimeSiam}_{\text{Base}}$ in the $\text{TSLD}_{\text{1G}}$ finetuning scenarios, especially in the Traffic benchmark. This observation highlights the efficacy of TimeSiam and the positive correlation between data-model size and the final performance.

\paragraph{Adapt to \update{Extended}-Length Input}

As illustrated in Eq.~\eqref{equ:extended}, TimeSiam can natively adapt to longer inputs. Figure \ref{fig:long_lookback} shows that the standard prediction framework may degenerate under extended input length, which may be because of the noises in longer series. Contrarily, benefiting from an ingenious integration of Siamese modeling and lineage embeddings, TimeSiam achieves more accurate predictions, even when predicting from extended input series. 

\begin{figure}[!h]
\begin{center}
    \centerline{\includegraphics[width=0.30\textwidth]{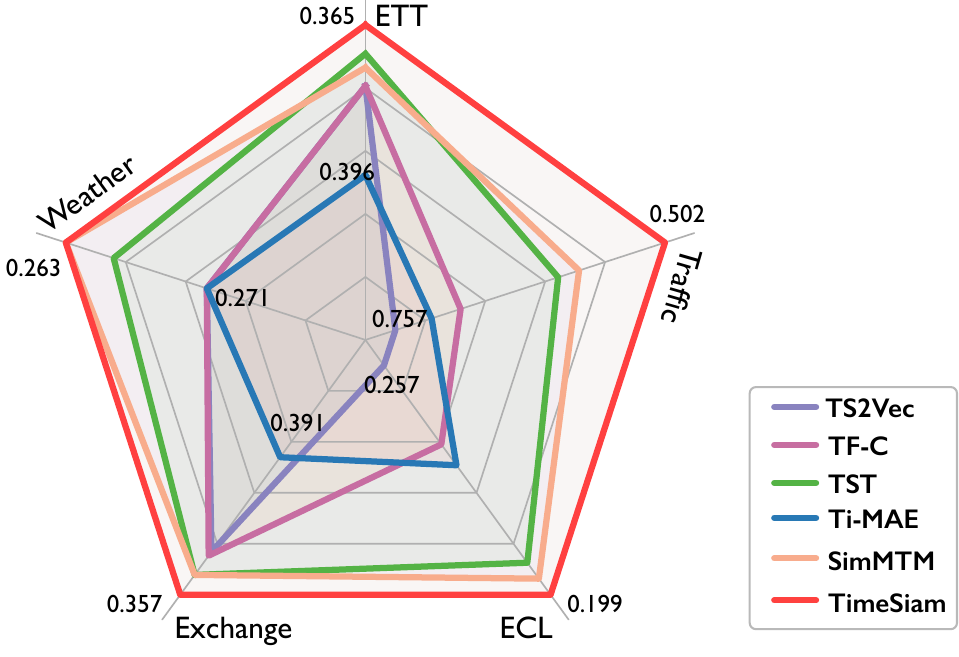}}
    \vspace{-10pt}
    \caption{Linear probing on in-domain forecasting setting. Average results (MSE) are reported. Full results are shown in Table \ref{tab:PatchTST_forecasting_linear_indomain_full}.}
    \label{fig:star}
\end{center}
\vspace{-25pt}
\end{figure}

\vspace{-5pt}
\paragraph{Linear Probing} As an important finetuning setting, we also experiment with the linear probing, where we fix the pre-trained encoder and only finetune the newly added projector at the end of the model. \camera{Figure \ref{fig:star} illustrates that TimeSiam demonstrates superior performance compared to other baselines in terms of overall \emph{linear probing} performance. Interestingly, by only fine-tuning the model head, the average forecasting performance across the four ETT subsets is already comparable with the results obtained through full fine-tuning, and significantly outperforms training from random initialization (\emph{MSE}: 0.365 vs. 0.383). This finding further validates the effectiveness of TimeSiam in learning generalizable representations for various downstream tasks.}

\vspace{-5pt}
\paragraph{Embedding Effectiveness}

\begin{figure}[h]
\begin{center}
    \centerline{\includegraphics[width=0.5\textwidth]{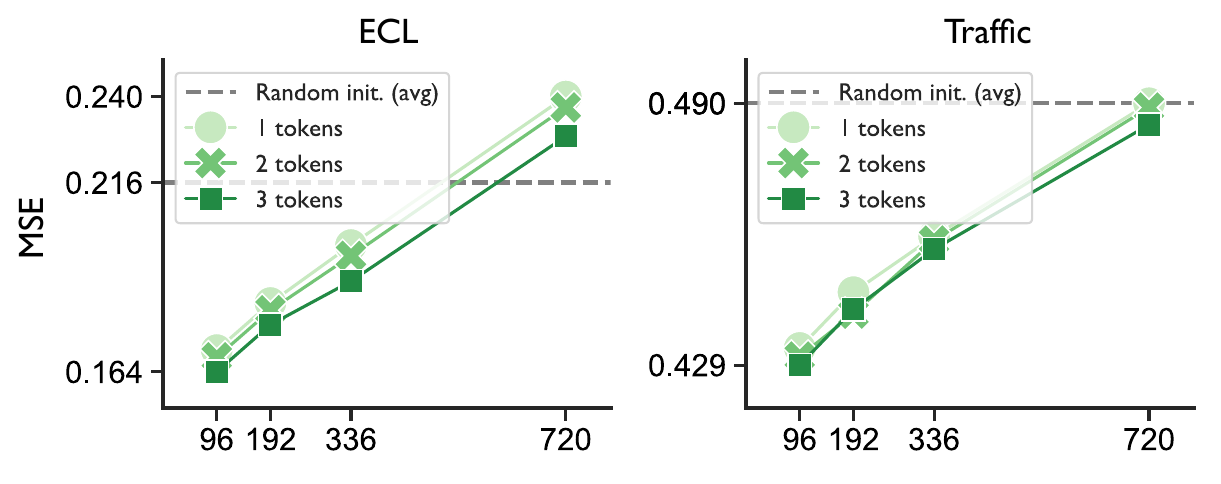}}
    \vspace{-10pt}
    \caption{Increasing number of lineage embeddings on the ECL and Traffic. All results are under the “input-96” in-domain setting.}
    \label{fig:tokens_effect}
    \vspace{-15pt}
\end{center}
\end{figure}

To elucidate the advantages of employing varying numbers of lineage embeddings within a fixed sampling range for prediction, as illustrated in Figure \ref{fig:tokens_effect}, our findings consistently demonstrate that the incorporation of lineage embeddings enhances prediction performance. Furthermore, augmenting the number of embeddings to encompass a greater extent of lineage within reasonable limits reinforces the efficacy of long-term prediction. Experimental results validate that lineage embeddings introduce more diverse temporal semantic information, enabling discrimination between different temporally distanced Siamese series, thereby boosting long-term prediction outcomes.

\section{Conclusion}

This paper proposes a simple, effective self-supervised pre-training framework named TimeSiam, uniquely focusing on temporal correlation modeling. TimeSiam employs Siamese networks as share encoders for randomly sampled past and current Siamese subseries. It further enhances data diversity through masking augmentation, which can also foster time-dependent representation learning by reconstructing current subseries from past observations. Additionally, we implement learnable lineage embeddings that efficiently capture disparities among Siamese subseries under different distances, enhancing the model's ability to cover diverse temporal correlations. Experimentally, TimeSiam demonstrated remarkable performance on various time series analysis tasks, consistently outperforming existing state-of-the-art baselines in both in- and cross-domain scenarios.

\section*{Acknowledgments}

\camera{This work was supported by the National Key Research and Development Plan (2021YFB1715200), the National Natural Science Foundation of China (62022050 and U2342217), the BNRist Innovation Fund (BNR2024RC01010), and the National Engineering Research Center for Big Data Software. We are grateful to our colleagues Yipeng Huang and Zhiyao Cen for their support in the experimental efficiency.}

\section*{Impact Statement}

This paper focus on developing practical time series pre-training methods. We presents a novel approach based on Siamese networks, which could provide some inspiration for future research. The experimental results demonstrate the effectiveness of our approach across various domains and its potential value for real-world applications. It is essential to note that our work focuses solely on scientific issues, and we also ensure that ethical considerations are carefully taken into account. All the medical-related datasets are publicly available for scientific research. Thus, we believe that there is no ethical risk associated with our research.

\nocite{langley00}

\bibliography{reference}
\bibliographystyle{icml2024}

\newpage
\appendix
\onecolumn

\section{Implementation Details}\label{app:implementation}

In this paper, all experiments were conducted on a single NVIDIA A100 SXM4 80GB GPU and implemented using the PyTorch framework~\cite{Paszke2019PyTorchAI} for five repetitions. We evaluated performance using Mean Square Error (MSE) and Mean Absolute Error (MAE) for time series forecasting. For classification tasks, we comprehensively assessed model performance by measuring accuracy, precision, recall, F1 score, AUROC, and AUPRC.

\subsection{Baseline Implementation}\label{app:baseline_implementation}

We have followed and compared the official implementations of all baselines to our approach. We have maintained the original configurations outlined in these papers to ensure a fair comparison. Note that we utilized an unofficial coding version of Ti-MAE \cite{li2023ti-TiMAE} due to the unavailability of its official open-source implementation.
 
 \begin{table}[h]
     \setlength{\abovecaptionskip}{0.cm}
    \setlength{\belowcaptionskip}{-0.cm}
  \label{tab:model_detail}
  \caption{Categories and open-source implementations of all baselines.}
  \vspace{5pt}
  \centering
  \renewcommand{\multirowsetup}{\centering}
  \begin{small}
  \setlength{\tabcolsep}{18pt}
  \renewcommand\arraystretch{2.0}
  \begin{tabular}{c|l|l}
    \toprule
    \textsc{Categories} & \textsc{Methods} & \textsc{Official Code Link} \\
    \midrule
    \multirow{5}{*}{Contrastive Learning} & \textsc{TS2Vec}~\cite{Yue2022-TS2Vec} & \href{https://github.com/yuezhihan/ts2vec}{https://github.com/yuezhihan/ts2vec} \\
    & \textsc{CoST}~\cite{woo2022cost-CoST} & \href{https://github.com/salesforce/CoST}{$\text{https://github.com/salesforce/CoST}$} \\
    & \textsc{LaST}~\cite{wang2022learning-LaST} & \href{https://github.com/zhycs/LaST}{https://github.com/zhycs/LaST} \\
    & \textsc{TF-C}~\cite{Zhang2022-TF-C} & \href{https://github.com/ mims-harvard/TFC-pretraining}{https://github.com/ mims-harvard/TFC-pretraining} \\
    & \textsc{COMET}~\cite{wang2023contrast} & \href{https://github.com/DL4mHealth/COMET}{https://github.com/DL4mHealth/COMET} \\
    \midrule
    \multirow{4}{*}{Masked Modeling} & \textsc{TST}~\cite{zerveas2021transformer-TST} & \href{https://github.com/gzerveas/mvts_transformer}{https://github.com/gzerveas/mvts\_transformer} \\
    & \textsc{Ti-MAE}~\cite{li2023ti-TiMAE} & \href{https://github.com/asmodaay/ti-mae}{https://github.com/asmodaay/ti-mae} \\
    & \textsc{PatchTST}~\cite{nie2022time-patch} & \href{https://github.com/yuqinie98/PatchTST}{https://github.com/yuqinie98/PatchTST} \\
    & \textsc{SimMTM}~\cite{xie2022simmim} & \href{https://github.com/thuml/simmtm}{https://github.com/thuml/simmtm} \\
    \bottomrule
  \end{tabular}
    \end{small}
\end{table}

\subsection{Training Configuration}

We construct two types of pre-training and fine-tuning scenarios, in-domain and cross-domain, based on benchmarks for prediction and classification tasks to compare the effectiveness of our method with other time series pre-training methods. In the pre-training phase, we pre-train the model with different learning rates and batch sizes based on the pre-trained dataset. We then fine-tune it for downstream prediction and classification tasks supervised by L2 and Cross-Entropy loss, respectively. The configuration details are in Table \ref{tab:model_detail}. Also, considering the size of the fine-tuned dataset and consistency with existing works, we fine-tune the model for 10 epochs for the prediction task and 50 epochs for the classification task. 

 \begin{table}[h]
     \setlength{\abovecaptionskip}{0.cm}
    \setlength{\belowcaptionskip}{-0.cm}
  \caption{Pre-training and fine-tuning configurations in forecasting and classification tasks.}
  \label{tab:model_detail}
  \vspace{5pt}
  \centering
  \renewcommand{\multirowsetup}{\centering}
  \setlength{\tabcolsep}{9pt}
  \begin{small}
  \renewcommand\arraystretch{2.0}
  \begin{tabular}{c|ccc|ccccccc}
    \toprule
    \multirow{2}{*}{\textsc{Tasks}} & \multicolumn{3}{c}{\textsc{Pre-training}} & \multicolumn{4}{c}{\textsc{Fine-tuning}} \\
    \cmidrule(lr){2-4}\cmidrule(lr){5-8}
     & learning rate  & batch size & epochs & learning rate & loss function & batch size & epochs \\
    \midrule
    Forecasting & $1\rm{e}\mbox{-}4$ & 32 & 50 & $1\rm{e}\mbox{-}4$ & L2 & \{8, 16, 32\} & 10 \\
    \midrule
    Classification & $1\rm{e}\mbox{-}4$ & 256 & 100 & $1\rm{e}\mbox{-}4$ & Cross-Entropy & \{32, 64, 128\} & 50 \\
    \bottomrule
  \end{tabular}
  \end{small}
\end{table}

\subsection{Model Configuration}\label{appdix:backbone}

We compare TimeSiam against eight state-of-the-art baselines for an unbiased and comprehensive comparison. To ensure the fairness of the evaluation, we choose state-of-the-art time series analysis models as a unified backbone for these pre-trained methods. Specifically, PatchTST \cite{nie2022time-patch} and iTransformer \cite{liu2023itransformer} are adopted for forecasting and employ Temporal Convolution Network (TCN)~\cite{bai2018tcn} for classification following the setup in \cite{wang2023contrast}. 

In addition, we performed a hyperparameter search for all baselines, adhering to their official configuration in the in-domain setting. For Siamese encoders, we explored various configurations by adjusting the number of encoder layers ($e_{\text{layers}}$) and decoder layers ($d_{\text{layers}}$) from $\{1, 2, 3, 4\}$, selecting hidden dimensions ($d_{\text{model}}$) from $\{16, 32, 64, 128, 256, 512\}$ and attention heads ($n_{\text{heads}}$) from $\{8, 16, 32\}$. In the case of TCN models, we investigated different numbers of residual blocks, considering configurations of $\{5, 8, 10\}$. During the fine-tuning stage, we carefully consider the learning rate (lr) from $\{1\rm{e}\mbox{-}3, 5\rm{e}\mbox{-}4, 1\rm{e}\mbox{-}4, 1\rm{e}\mbox{-}5\}$, and head dropout (dropout) from $\{0, 0.1, 0.2, 0.3\}$ in order to enhance the adaptability of our pretrained model to diverse datasets.

Primarily, two model configurations with different sizes are explored in the cross-domain forecasting setting, that is \textbf{TimeSiam-${\text{Base}}$} and \textbf{TimeSiam-${\text{Large}}$}. These two models are used to evaluate the impact of model capacity on forecasting performance, specifically in the context of cross-domain pre-training and fine-tuning on large-scale data.

 \begin{table}[h]
     \setlength{\abovecaptionskip}{0.cm}
    \setlength{\belowcaptionskip}{-0.cm}
  \label{tab:model_detail}
  \caption{Two experimental configurations of TimeSiam with different model sizes.}
  \vspace{5pt}
  \centering
  \renewcommand{\multirowsetup}{\centering}
  \setlength{\tabcolsep}{20pt}
  \begin{small}
  \renewcommand\arraystretch{2.0}
  \begin{tabular}{cccccccc}
    \toprule
     \multirow{2}{*}{\textsc{Types}} & \multicolumn{5}{c}{\textsc{Configuration}}  & \multirow{2}{*}{\textsc{Parameters}} \\
     \cmidrule(lr){2-6}
     & $e_{\text{layers}}$ & $d_{\text{layers}}$ & $d_{\text{model}}$ & $d_{\text{ff}}$  & $n_{\text{heads}}$ & \\
    \midrule
    TimeSiam$_{\text{Base}}$ &  3 & 1 & 128 & 256 & 8 & 709,344 \\
    \midrule
    TimeSiam$_{\text{Large}}$ & 5 & 2 & 128 & 1024 & 16 & 2,554,720 \\
    \bottomrule
  \end{tabular}
  \end{small}
  \vspace{10pt}
\end{table}

\section{Dataset Description}\label{app:datasets}

We conduct experiments on eleven well-established datasets and two newly constructed datasets covering two primary tasks in time series analysis: forecasting and classification. These datasets cover a variety of application scenarios, different types of signals, multivariate channel dimensions, varying time series lengths, large-span sampling frequencies, and different data sizes. The detailed descriptions of these datasets are summarized in Table \ref{tab:dataset_desc}.

 \begin{table}[h]
  \caption{Dataset descriptions. \emph{Samples} are organized in (Train/Validation/Test).}
  \vspace{5pt}
  \label{tab:dataset_desc}
  \centering
  \renewcommand{\multirowsetup}{\centering}
  \setlength{\tabcolsep}{3.5pt}
  \begin{small}
  \renewcommand\arraystretch{2.0}
  \begin{tabular}{c|c|c|c|c|c|c|c}
    \toprule
    \scalebox{1.0}{\textsc{Tasks}} & \scalebox{1.0}{\textsc{Datasets}} & \scalebox{1.0}{\textsc{Channels}} & \scalebox{1.0}{\textsc{Series Length}} & \scalebox{1.0}{\textsc{Samples}} & \scalebox{1.0}{\textsc{Classes}} & \scalebox{1.0}{\textsc{Information}} & \scalebox{1.0}{\textsc{Frequency}} \\
    \toprule
    \multirow{8}{*}{\rotatebox{90}{\scalebox{1.0}{Forecasting}}} & \scalebox{1.0}{ETTh1,ETTh2} & 7 & \scalebox{1.0}{\{96,192,336,720\}} & \scalebox{1.0}{8,545/2,881/2,881} & - & \scalebox{1.0}{Electricity} & \scalebox{1.0}{Hourly} \\
    & \scalebox{1.0}{ETTm1,ETTm2} & \scalebox{1.0}{7} & \scalebox{1.0}{\{96,192,336,720\}} & \scalebox{1.0}{34,465/11,521/11,521} & - & \scalebox{1.0}{Electricity} & \scalebox{1.0}{15 Mins} \\
    & \scalebox{1.0}{Weather} & \scalebox{1.0}{21} & \scalebox{1.0}{\{96,192,336,720\}} & \scalebox{1.0}{36,792/5,271/10,540} & - & \scalebox{1.0}{Weather} & \scalebox{1.0}{10 Mins} \\
    & \scalebox{1.0}{Exchange} & \scalebox{1.0}{8} & \scalebox{1.0}{\{96,192,336,720\}} & \scalebox{1.0}{5,120/665/1,422} & - & \scalebox{1.0}{Exchange rate} & \scalebox{1.0}{Daily} \\
    & \scalebox{1.0}{Electricity} & \scalebox{1.0}{321} & \scalebox{1.0}{\{96,192,336,720\}} & \scalebox{1.0}{18,317/2,633/5261} & - & \scalebox{1.0}{Electricity} & \scalebox{1.0}{Hourly} \\
    & \scalebox{1.0}{Traffic} & \scalebox{1.0}{862} & \scalebox{1.0}{\{96,192,336,720\}} & \scalebox{1.0}{12,185/1,757/3,509} & - & \scalebox{1.0}{Transportation} & \scalebox{1.0}{Hourly} \\
    & \scalebox{1.0}{TSLD-500M} & \scalebox{1.0}{1} & \scalebox{1.0}{\{96,192,336,720\}} & \scalebox{1.0}{369,030/31,872/-} & - & \scalebox{1.0}{Multi-domain} & \scalebox{1.0}{Mixing} \\
    & \scalebox{1.0}{TSLD-1G} & \scalebox{1.0}{1} & \scalebox{1.0}{\{96,192,336,720\}} & \scalebox{1.0}{13,984,175/1,061,806/-} & - & \scalebox{1.0}{Multi-domain} & \scalebox{1.0}{Mixing} \\
    \midrule
    \multirow{3}{*}{\rotatebox{90}{\scalebox{1.0}{Classification}}} & \scalebox{1.0}{AD} & \scalebox{1.0}{16} & \scalebox{1.0}{256} & \scalebox{1.0}{4,329/891/747} & \scalebox{1.0}{3} & \scalebox{1.0}{EEG} & \scalebox{1.0}{256 Hz} \\
    & \scalebox{1.0}{TDBrain} & \scalebox{1.0}{33} & \scalebox{1.0}{256} & \scalebox{1.0}{8,208/1,824/1,824} & \scalebox{1.0}{3} & \scalebox{1.0}{EEG} & \scalebox{1.0}{500 Hz} \\
    & \scalebox{1.0}{PTB} & \scalebox{1.0}{15} & \scalebox{1.0}{300} & \scalebox{1.0}{53,950/3,400/5,020} & \scalebox{1.0}{3} & \scalebox{1.0}{ECG} & \scalebox{1.0}{1000 Hz} \\
    \bottomrule
  \end{tabular}
    \end{small}
\end{table}

\subsection{Forecasting Datasets}

(1) \textbf{ETT (4 subsets)} ~\cite{zhou2021informer-ETT} contains a group of four subsets oil temperature and power load collected by electricity transformers from July 2016 to July 2018 with minutes or hourly recorded frequence. 

(2) \textbf{Weather} ~\cite{weatherdata} includes meteorological time series with 21 weather indicators collected every 10 minutes from the Weather Station of the Max Planck Biogeochemistry Institute in 2020.

(3) \textbf{Electricity} ~\cite{ecldata} records the hourly electricity consumption of 321 clients from 2012 to 2014. 

(4) \textbf{Traffic} ~\cite{trafficdata} encompasses the hourly measures of road occupancy rates obtained from 862 sensors situated in the San Francisco Bay area freeways between January 2015 and December 2016. 

(5) \textbf{Exchange} ~\cite{lai2018exchange} records the daily exchange rates of eight different countries ranging from 1990 to 2016.

\subsection{Classification Datasets}

(1) \textbf{AD} ~\cite{escudero2006ad} has electroencephalography (EEG) recordings from 12 Alzheimer's patients and 11 healthy controls. Each patient has around 30 trials, each lasting for 5 seconds with 1280 timestamps (sampled at 256Hz) and includes 16 channels. 

(2) \textbf{PTB} ~\cite{goldberger2000ptb} has electrocardiogram (ECG) recordings from 290 patients with 15 channels sampled at 1000 Hz. This paper focuses on a subset of the dataset that includes 198 patients with heart diseases: Myocardial infarction and healthy controls.

(3) \textbf{TDBrain} ~\cite{van2022tdbrain} monitors brain signals of 1274 patients with 33 channels during  EC (Eye closed) and EO (Eye open) tasks. It includes 60 types of diseases, but this paper focuses on a subset of 25 Parkinson's disease patients and 25 healthy controls. Only the EC task trials are used for representation learning.

\subsection{Merged Large Scale Datasets}

To further substantiate the significance of time series pre-training on large-scale data and showcase its benefits in diverse and extensive datasets, we have amalgamated multiple non-overlapping time series datasets from various domains to construct the \textbf{T}ime \textbf{S}eries \textbf{L}arge \textbf{D}atasets (\textbf{TSLD}). In this paper, we present two versions of TSLD to valid our approach.

\vspace{5pt}
(1) \textbf{TSLD-500M} is a composite dataset comprising 400,902 samples from 12 time series datasets across the domains of Electricity, Transport, Energy, Climate, and others.

\vspace{5pt}
(2) \textbf{TSLD-1G}, building upon the TSLD-500M dataset, incorporates additional diverse datasets from domains such as Society, IoT, and Web. With an impressive sample count of 15,045,981 observations, TSLD-1G surpasses the size of datasets commonly used in time series analysis and provides greater diversity.

\section{Masking Strategy}\label{app:masking}

\begin{figure}[t]
\begin{center}
    \centerline{\includegraphics[width=\textwidth]{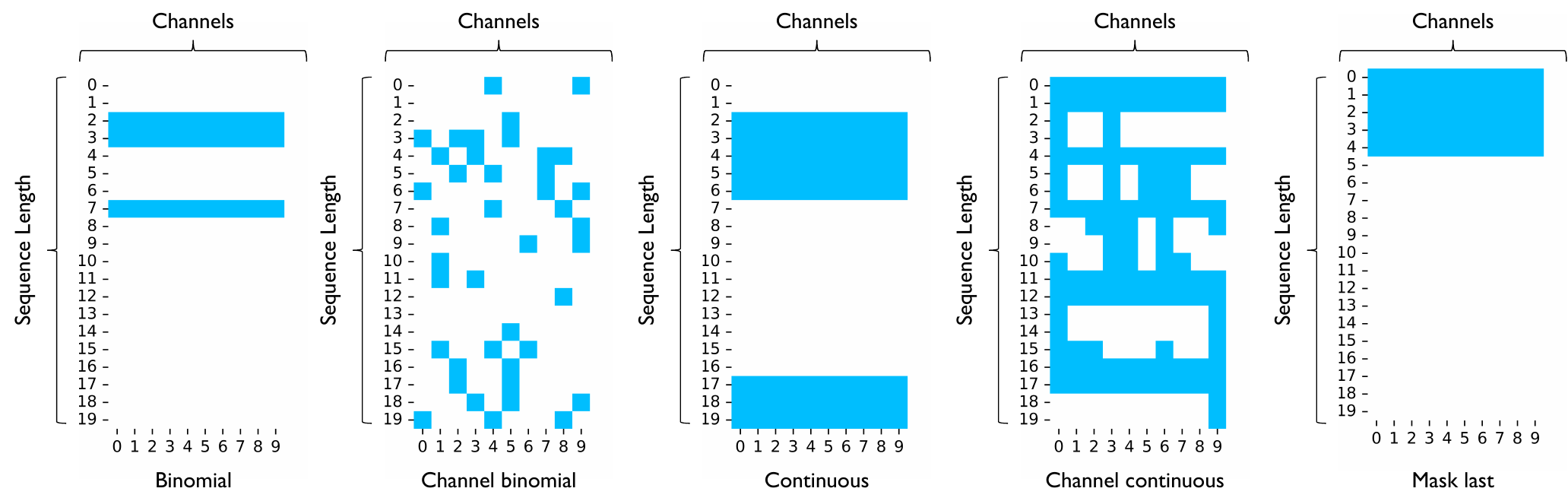}}
    \vspace{-10pt}
    \caption{Showcases of various masking rules (75\% masked ratio). The x-axis shows the channels and the y-axis represents the sequence length of the time series. Blue blocks indicate unmasked time stamps while white blocks represent masked ones.}
    \label{fig:masking}
    \vspace{-20pt}
\end{center}
\end{figure}

In this paper, we explored five different mask rules: binomial, channel binomial, continuous, channel continuous, and only masking the last to assess their impact on TimeSiam, illustrated in Figure \ref{fig:masking}.

(1) \textbf{Binomial masking}: Generate a mask by employing a binomial distribution across all channels within a given sample.

(2) \textbf{Channel binomial masking}: Generate a mask based on a binomial distribution that selectively masks individual channels at different timestamps within the sample.

(3) \textbf{Continuous masking}: Generate a mask by employing a geometric distribution across all channels within a given sample.

(4) \textbf{Channel continuous masking}: Generate a mask based on a geometric distribution that selectively masks individual channels at different timestamps within the sample.

(5) \textbf{Masking last}: Only mask the tail of time series in all channels.

\section{Linear Probing and Full Fine-tuning}

\begin{figure}[h]
\begin{center}
    \centerline{\includegraphics[width=\textwidth]{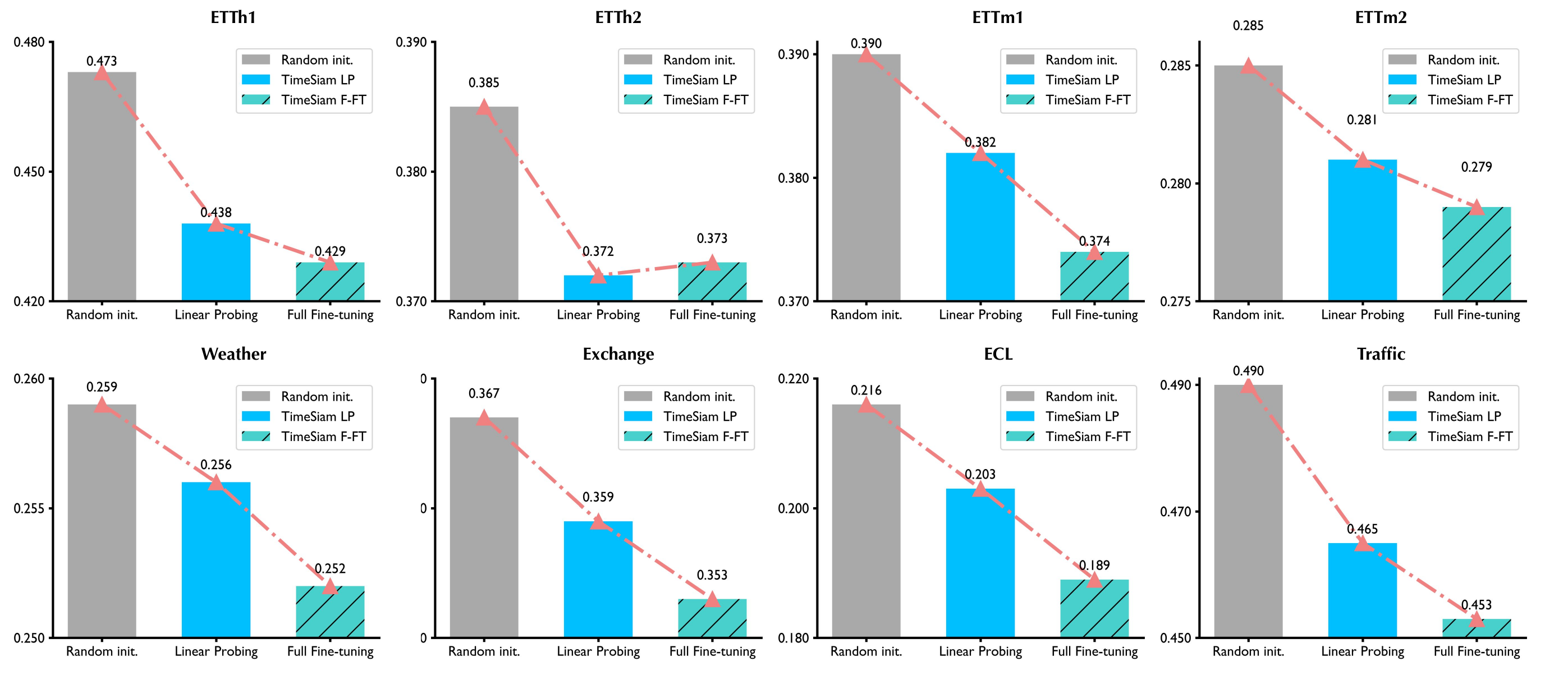}}
    \caption{Comparison is made between the performance of linear probing pre-trained from TSLD-1G on various datasets and pre-training followed by fine-tuning on the same dataset. The mean squared error (MSE) is computed across all prediction lengths and serves as a measure of performance. A lower averaged MSE indicates superior predictive capability.}
    \label{fig:linear_probing}
\end{center}
\end{figure}

The results depicted in Figure \ref{fig:linear_probing} unequivocally demonstrate that both fine-tuning and linear probing methodologies utilizing TimeSiam outperform fully supervised learning from random initiation. Moreover, the findings suggest that full fine-tuning consistently yields superior results compared to linear probing across most datasets, with ETTh2 being a notable exception, where both approaches exhibit comparable performance.

\section{Multiple Lineages Representation Visualization}

We employ Principal Components Analysis (PCA) to elucidate the distribution of temporal representations on the ECL dataset.  \camera{We will only train the learned lineage embeddings during the pre-training phase. However, during downstream fine-tuning or linear probing, we will keep them fixed and not update them. It is worth noting that the embedded feature will be the same without different lineage embeddings. However,} when time series is fed into a pre-trained Siamese network with different lineage embeddings, the model generates divergent temporal representations that representations derived from the same lineage embeddings tend to be closely clustered together, while representations from different lineage embeddings exhibit significant dissimilarity. \camera{Upon visual analysis, we have observed that the representations generated based on the same data but with different lineage embeddings exhibit a high level of diversity. This observation effectively validates the effectiveness of combining a pre-trained Siamese network with different lineage embeddings, which can enlarge the representation diversity.}


\begin{figure*}[h]
\begin{center}
    \centerline{\includegraphics[width=0.7\textwidth]{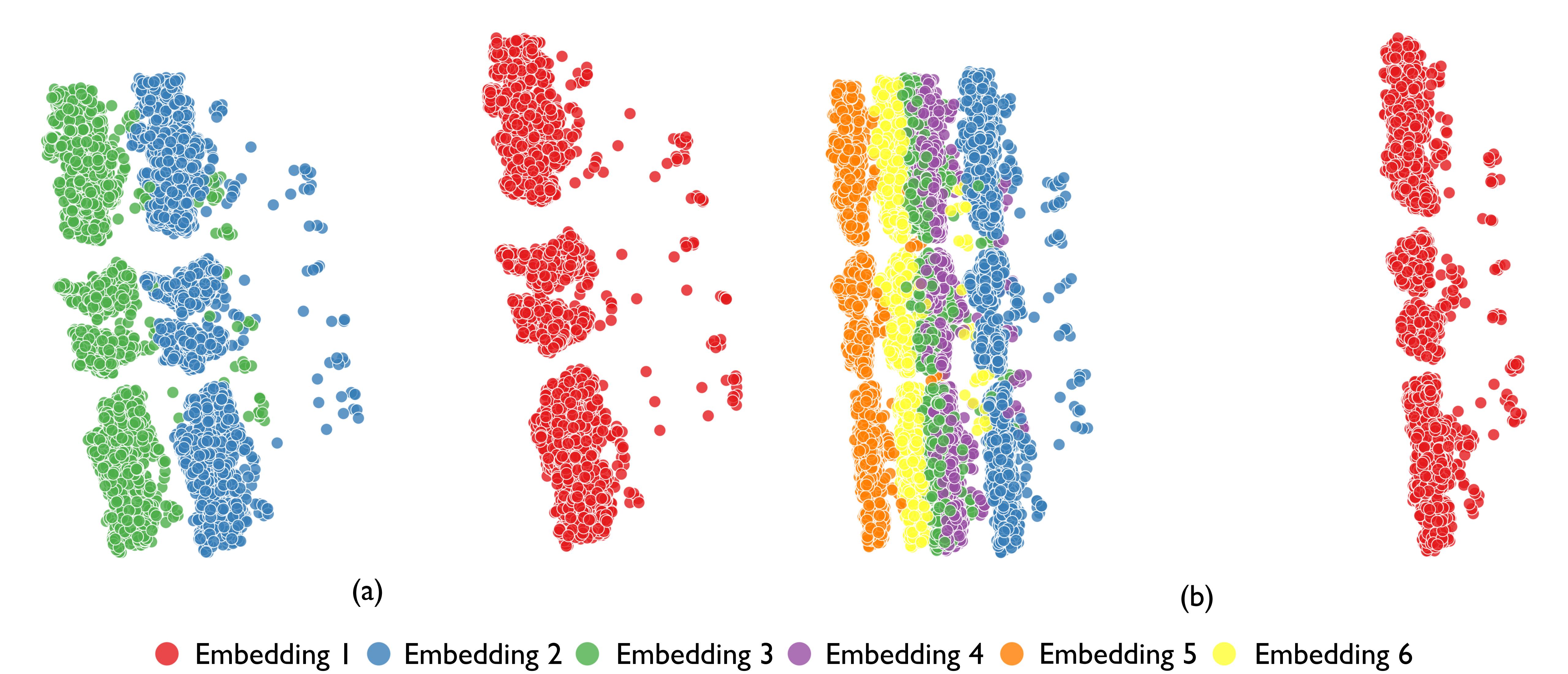}}
    \caption{Visualizing the effect of temporal shift representations. (a) Visualization of test distribution under three types of lineage embeddings for ECL. (b) Visualization of test distribution under six types of lineage embeddings for ECL.}
\end{center}
\vspace{-20pt}
\end{figure*}

\section{Adapt Extended Input Length} \label{app:adapt_length_full_result}

 To facilitate the performance of Timesiam on fine-tuning scenarios with extended input lengths, we choose the input length to be an integral multiple of the pre-training length. In practice, the series length is not restricted to be an integral multiple of the pre-training length. TimeSiam can handle flexible input lengths, as different lineage embeddings can be shared across different time segments.

\begin{figure*}[!hb]
\begin{center}
    \centerline{\includegraphics[width=0.6\textwidth]{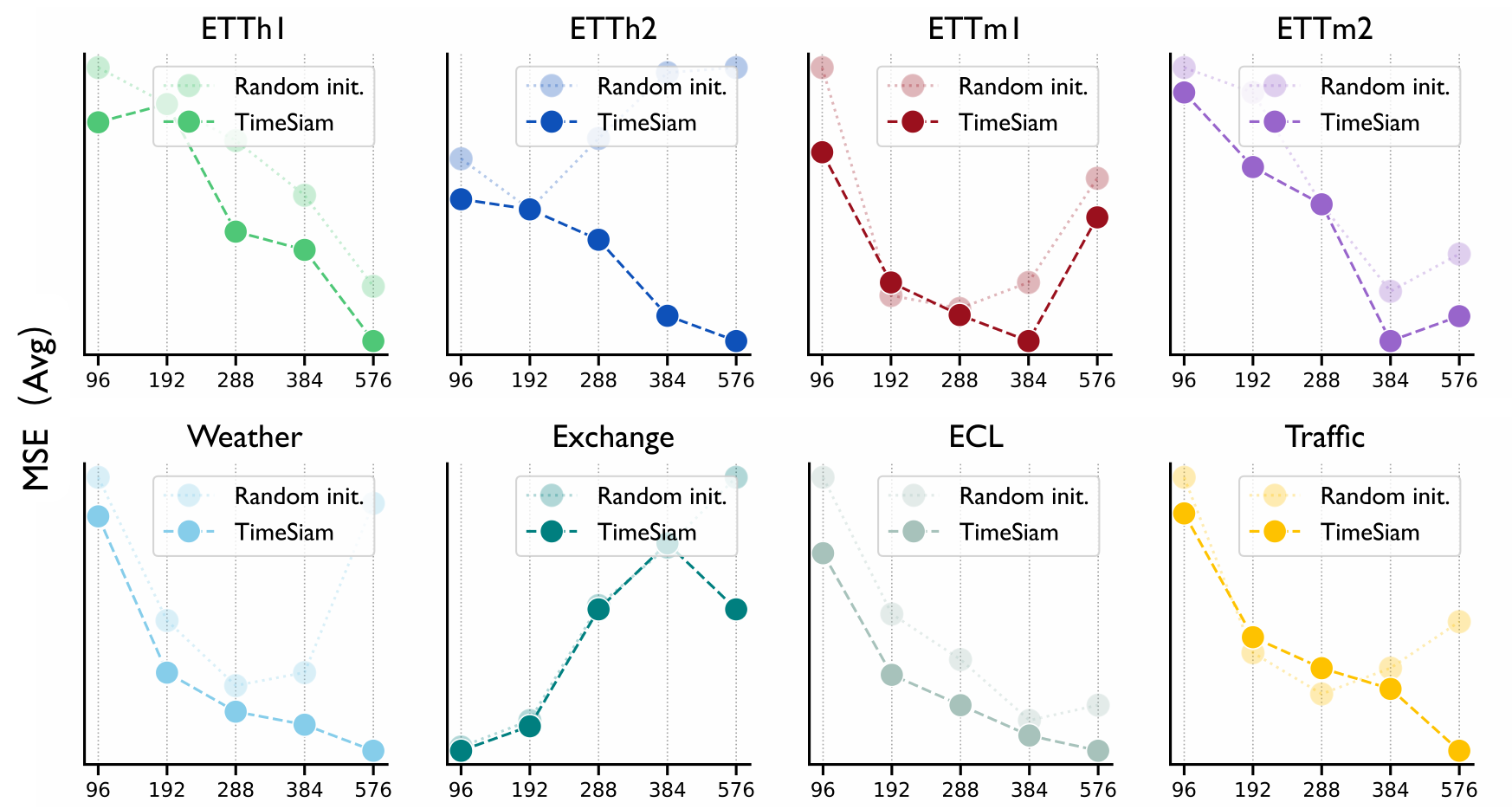}}
    \vspace{-5pt}
    \caption{Full results for fine-tuning the pre-trained model with extended input length, where the input length is selected from \{96, 192, 288, 384, 576\}. The MSE averaged from four future lengths $O \in \{96,192,336,720\}$ is reported.}
\end{center}
\end{figure*}

\section{Full Results}\label{app:full results}

Due to the limited length of the text, we summarize the main experiments as follows:

 \begin{table}[h] \label{app:mainresults_idx}
    \vspace{-10pt}
  \caption{The main results for time series forecasting and classification tasks.}
  \vspace{5pt}
  \label{tab:mainresults_idx}
  \centering
  \begin{small}
  \renewcommand{\multirowsetup}{\centering}
  \setlength{\tabcolsep}{30pt}
  \renewcommand\arraystretch{2.0}
  \begin{tabular}{c|c|c|c}
    \toprule
    \textsc{Experiments Categories} & \textsc{\scalebox{1.0}{Tasks}} & \textsc{\scalebox{1.0}{Evaluation}} & \textsc{\scalebox{1.0}{Tabels Name}} \\
    \toprule
    \multirow{4}{*}{The main experiment} & \multirow{2}{*}{Forecasting} & \multirow{1}{*}{\scalebox{1.0}{In-domain}} & \scalebox{1.0}{Table \ref{tab:PatchTST_forecasting_indomain_full}, \ref{tab:iTransformer_forecasting_indomain_full}} \\
    & & \multirow{1}{*}{\scalebox{1.0}{Cross-domain}} & \scalebox{1.0}{Table \ref{tab:PatchTST_forecasting_crossdomain_full}} \\
    \cmidrule(lr){2-4}
    & \multirow{2}{*}{Classification} & \multirow{1}{*}{\scalebox{1.0}{In-domain}} & \scalebox{1.0}{Table \ref{tab:classification_indomain_full}} \\
    & & \multirow{1}{*}{\scalebox{1.0}{Cross-domain}} & \scalebox{1.0}{Table \ref{tab:classification_crossdomain_full}} \\
    \bottomrule
  \end{tabular}
  \end{small}
\end{table}

\vspace{20pt}
\section{ShowCases} \label{app:showcases}

\subsection{Different Masked Ratios}

\begin{figure}[!htbp]
\begin{center}
    \centerline{\includegraphics[width=1.0\textwidth]{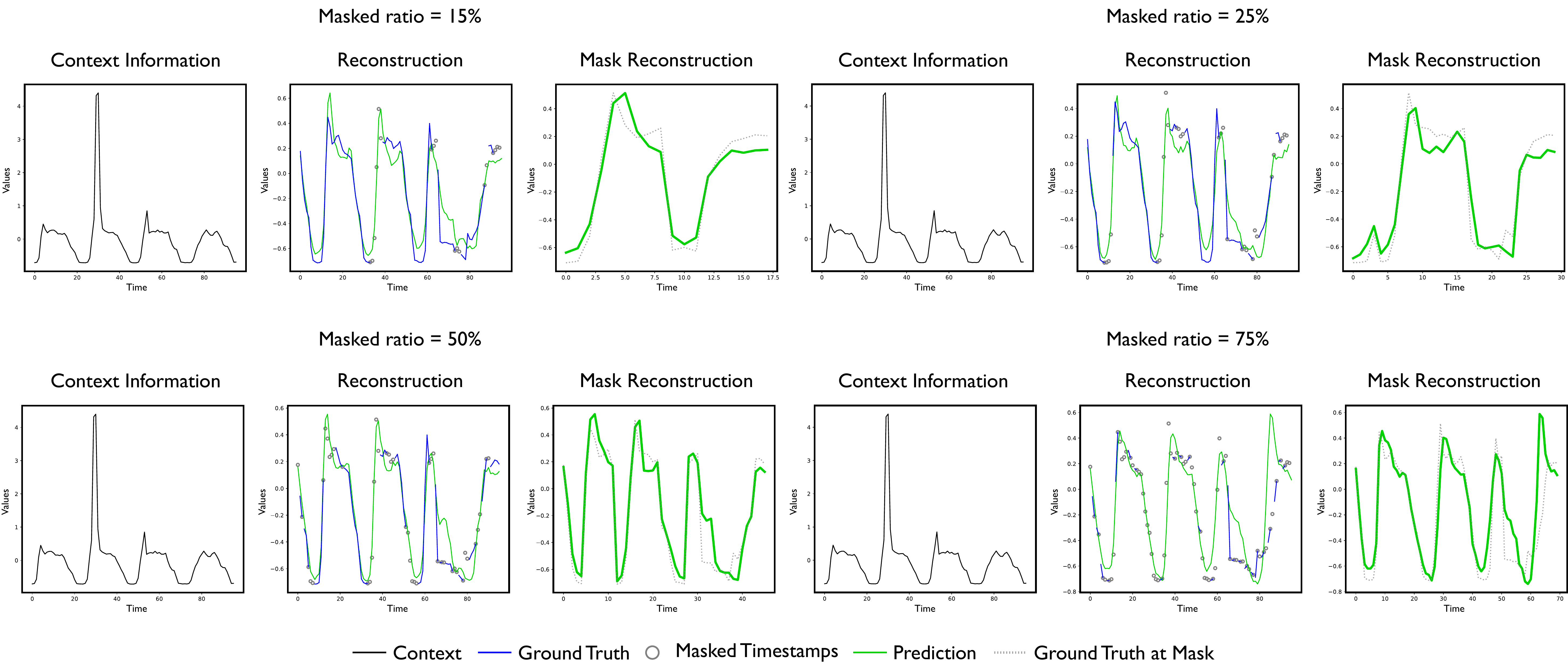}}
    \vspace{-5pt}
 	\caption{Showcases of TimeSiam in reconstructing time series with different masked ratios from Traffic.}
	\label{fig:show_case_masked_ratio}
\end{center}
\end{figure}

To investigate the reconstruction process of TimeSiam, we visually represent past time series, masked current time series, and reconstructed current time series with varying mask ratios using validation data from diverse datasets. Figure \ref{fig:show_case_masked_ratio} demonstrates the reconstruction effects of TimeSiam at different mask ratios applied to the current time series. The context information is obtained by random sampling based on the current series, and the reconstruction becomes more challenging as the mask ratio increases due to the limited available information. Nevertheless, our TimeSiam model consistently achieves accurate reconstruction of masked current time series despite the scarcity of data and significant variation in temporal dimension between past and present. This accomplishment highlights the effectiveness of our approach in learning internal time-dependent representations through a past-to-current reconstruction.

\subsection{Different Datasets}

We further demonstrate the reconstruction effect across various datasets with different data distributions, as detailed below.

\begin{figure}[!htbp]
\begin{center}
    \centerline{\includegraphics[width=1.0\textwidth]{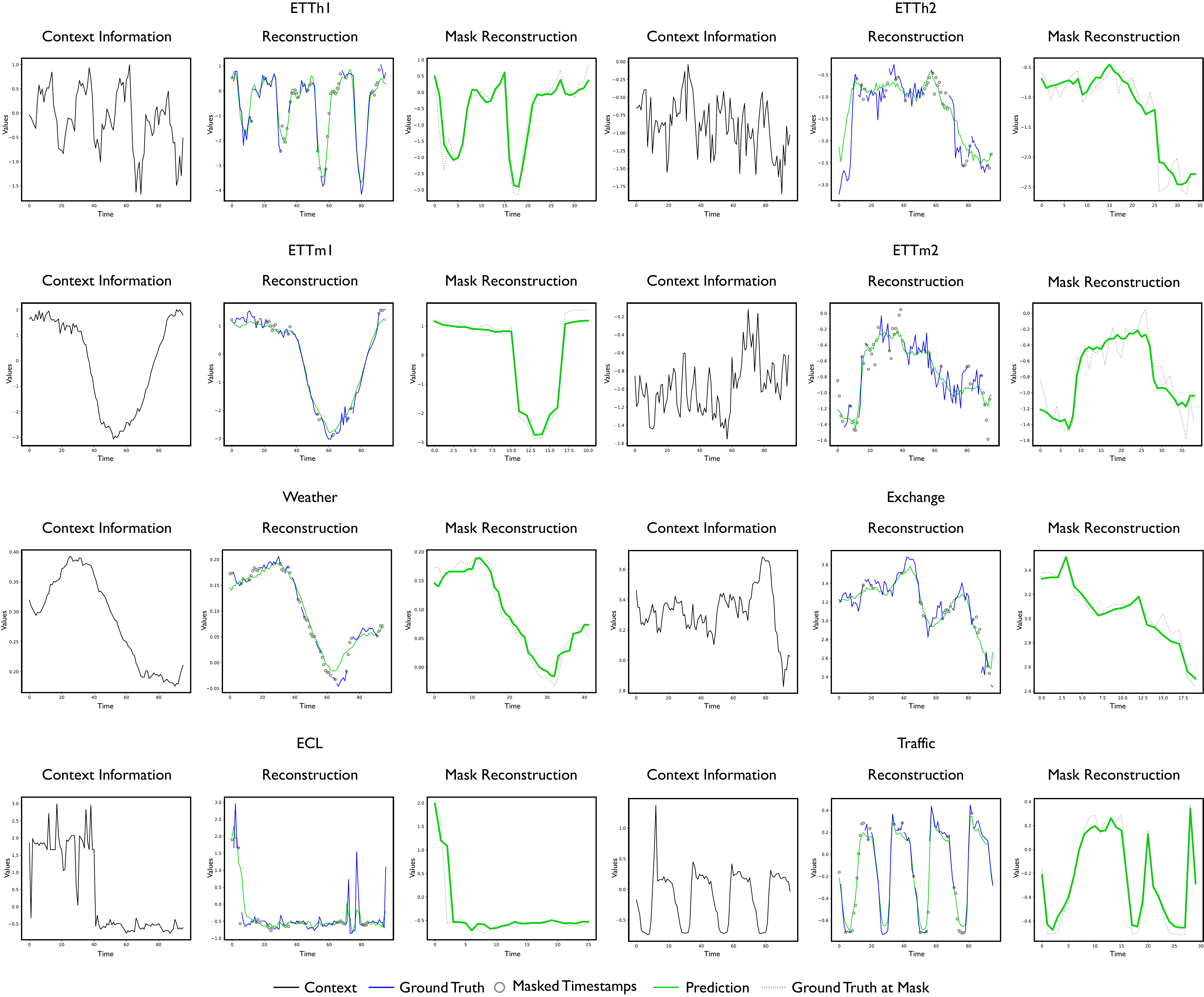}}
    \vspace{-5pt}
 	\caption{Showcases of TimeSiam in reconstructing time series from different datasets with 25\% masked raito.}
	\label{fig:show_case}
\end{center}
\end{figure}

\newpage

\begin{table*}[h]
  \caption{Full results for the in-domain setting of forecasting using PatchTST. Pre-training and fine-tuning are performed on the same datasets. The standard deviations are within 0.005 for MSE and within 0.004 for MAE.}
  \label{tab:PatchTST_forecasting_indomain_full}
  \vspace{-5pt}
  \vskip 0.15in
  \centering
  \resizebox{1.03\textwidth}{!}{
  \begin{small}
  \begin{sc}
  \renewcommand{\multirowsetup}{\centering}
  \setlength{\tabcolsep}{1.5pt}
  \renewcommand\arraystretch{1.0}
  \scalebox{0.9}{\begin{tabular}{c|c|cc|cccccccccccccccccc|cc}
    \toprule
    \multicolumn{2}{c}{Methods}  & \multicolumn{2}{c}{Random init.} & \multicolumn{2}{c}{CPC}  & \multicolumn{2}{c}{TNC} & \multicolumn{2}{c}{TS2Vec} & \multicolumn{2}{c}{CoST} & \multicolumn{2}{c}{LaST} & \multicolumn{2}{c}{TFC} & \multicolumn{2}{c}{TST} & \multicolumn{2}{c}{Ti-MAE} & \multicolumn{2}{c}{SimMTM} & \multicolumn{2}{c}{{\textbf{TimeSiame}}}\\
    \multicolumn{2}{c}{Metric} & mse & mae & mse & mae & mse & mae & mse & mae & mse & mae & mse & mae & mse & mae & mse & mae & mse & mae & mse & mae & mse & mae \\
    \toprule
    \multirow{5}{*}{\rotatebox{90}{ETTh1}}
    &  96 & 0.420 & 0.423 & 0.380 & 0.401 & 0.377 & 0.397 & 0.381 & 0.400 & 0.383 & 0.405 & 0.396 & 0.413 & 0.399 & 0.420 & 0.377 & 0.401 & 0.396 & 0.415 & 0.367 & 0.389 & 0.378 & 0.401 \\
    & 192 & 0.465 & 0.449 & 0.426 & 0.429 & 0.423 & 0.427 & 0.421 & 0.427 & 0.434 & 0.437 & 0.457 & 0.451 & 0.444 & 0.449 & 0.432 & 0.436 & 0.440 & 0.443 & 0.424 & 0.423 & 0.422 & 0.430 \\
    & 336 & 0.504 & 0.470 & 0.465 & 0.451 & 0.471 & 0.453 & 0.468 & 0.452 & 0.474 & 0.460 & 0.507 & 0.478 & 0.479 & 0.467 & 0.475 & 0.461 & 0.481 & 0.462 & 0.473 & 0.456 & 0.459 & 0.452 \\
    & 720 & 0.502 & 0.492 & 0.488 & 0.479 & 0.508 & 0.485 & 0.553 & 0.507 & 0.535 & 0.509 & 0.516 & 0.508 & 0.491 & 0.490 & 0.525 & 0.500 & 0.475 & 0.481 & 0.494 & 0.493 & 0.459 & 0.466 \\
    \cmidrule(lr){2-24}
    & Avg & 0.473 & 0.458 & 0.440 & 0.440 & 0.445 & 0.441 & 0.456 & 0.447 & 0.457 & 0.453 & 0.469 & 0.463 & 0.453 & 0.457 & 0.452 & 0.450 & 0.448 & 0.450 & 0.440 & 0.440 & \textbf{0.429} & \textbf{0.437} \\
    \midrule
    \multirow{5}{*}{\rotatebox{90}{ETTh2}}  
    &  96 & 0.297 & 0.345 & 0.313 & 0.364 & 0.291 & 0.343 & 0.297 & 0.343 & 0.288 & 0.342 & 0.294 & 0.345 & 0.302 & 0.345 & 0.304 & 0.358 & 0.300 & 0.353 & 0.299 & 0.352 & 0.293 & 0.345 \\
    & 192 & 0.388 & 0.400 & 0.392 & 0.412 & 0.366 & 0.393 & 0.366 & 0.392 & 0.367 & 0.392 & 0.379 & 0.395 & 0.369 & 0.392 & 0.379 & 0.403 & 0.372 & 0.396 & 0.380 & 0.398 & 0.370 & 0.392 \\
    & 336 & 0.426 & 0.434 & 0.438 & 0.449 & 0.427 & 0.438 & 0.416 & 0.430 & 0.413 & 0.429 & 0.423 & 0.436 & 0.412 & 0.428 & 0.412 & 0.432 & 0.418 & 0.431 & 0.422 & 0.432 & 0.410 & 0.424 \\
    & 720 & 0.431 & 0.446 & 0.460 & 0.470 & 0.433 & 0.451 & 0.424 & 0.447 & 0.428 & 0.446 & 0.445 & 0.460 & 0.428 & 0.446 & 0.438 &0.457  & 0.427 & 0.447 & 0.428 & 0.449 & 0.418 & 0.440 \\
    \cmidrule(lr){2-24}
    & Avg & 0.385 & 0.406 & 0.401 & 0.424 & 0.379 & 0.406 & 0.376 & 0.403 & 0.374 & 0.402 & 0.385 & 0.409 & 0.378 & 0.403 & 0.383 & 0.413 & 0.379 & 0.407 & 0.382 & 0.408 & \textbf{0.373} & \textbf{0.400} \\
    \midrule
    \multirow{5}{*}{\rotatebox{90}{ETTm1}} 
    &  96 & 0.330 & 0.368 & 0.324 & 0.361 & 0.323 & 0.361 & 0.325 & 0.364 & 0.348 & 0.377 & 0.345 & 0.381 & 0.353 & 0.378 & 0.319 & 0.360 & 0.325 & 0.363 & 0.317 & 0.356 & 0.319 & 0.360 \\
    & 192 & 0.369 & 0.385 & 0.368 & 0.382 & 0.366 & 0.383 & 0.370 & 0.389 & 0.367 & 0.387 & 0.372 & 0.391 & 0.361 & 0.384 & 0.360 & 0.387 & 0.363 & 0.385 & 0.362 & 0.387 & 0.353 & 0.379 \\
    & 336 & 0.400 & 0.407 & 0.403 & 0.405 & 0.399 & 0.405 & 0.405 & 0.415 & 0.404 & 0.414 & 0.412 & 0.420 & 0.392 & 0.406 & 0.391 & 0.408 & 0.396 & 0.409 & 0.387 & 0.405 & 0.383 & 0.402 \\
    & 720 & 0.460 & 0.439 & 0.461 & 0.439 & 0.457 & 0.439 & 0.471 & 0.452 & 0.460 & 0.447 & 0.462 & 0.448 & 0.448 & 0.440 & 0.449 & 0.445 & 0.452 & 0.438 & 0.443 & 0.438 & 0.440 & 0.436 \\
    \cmidrule(lr){2-24}
    & Avg & 0.390 & 0.400 & 0.389 & 0.397 & 0.386 & 0.397 & 0.393 & 0.405 & 0.395 & 0.406 & 0.398 & 0.410 & 0.389 & 0.402 & 0.380 & 0.400 & 0.384 & 0.399 & 0.377 & 0.397 & \textbf{0.374} & \textbf{0.394} \\
    \midrule
    \multirow{5}{*}{\rotatebox{90}{ETTm2}} 
    &  96 & 0.175 & 0.258 & 0.196 & 0.281 & 0.187 & 0.261 & 0.174 & 0.261 & 0.181 & 0.269 & 0.177 & 0.258 & 0.281 & 0.327 & 0.181 & 0.265 & 0.175 & 0.261 & 0.175 & 0.262 & 0.175 & 0.261 \\
    & 192 & 0.247 & 0.307 & 0.261 & 0.323 & 0.241 & 0.302 & 0.247 & 0.306 & 0.247 & 0.312 & 0.252 & 0.309 & 0.241 & 0.302 & 0.247 & 0.309 & 0.241 & 0.303 & 0.244 & 0.307 & 0.241 & 0.303 \\
    & 336 & 0.309 & 0.345 & 0.302 & 0.343 & 0.313 & 0.363 & 0.306 & 0.345 & 0.309 & 0.348 & 0.307 & 0.344 & 0.304 & 0.343 & 0.314 & 0.354 & 0.301 & 0.341 & 0.312 & 0.351 & 0.300 & 0.341 \\
    & 720 & 0.408 & 0.403 & 0.399 & 0.397 & 0.408 & 0.407 & 0.427 & 0.415 & 0.408 & 0.406 & 0.404 & 0.402 & 0.404 & 0.403 & 0.408 & 0.407 & 0.398 & 0.397 & 0.410 & 0.408 & 0.399 & 0.398 \\
    \cmidrule(lr){2-24}
    & Avg & 0.285 & 0.328 & 0.290 & 0.336 & 0.287 & 0.333 & 0.289 & 0.332 & 0.286 & 0.334 & 0.285 & 0.328 & 0.281 & 0.327 & 0.288 & 0.334 & 0.279 & 0.326 & 0.285 & 0.332 & \textbf{0.279} & \textbf{0.326} \\
    \midrule
    \multirow{5}{*}{\rotatebox{90}{Weather}} 
    &  96 & 0.177 & 0.218 & 0.193 & 0.230 & 0.191 & 0.231 & 0.174 & 0.216 & 0.171 & 0.214 & 0.170 & 0.212 & 0.177 & 0.218 & 0.177 & 0.221 & 0.175 & 0.218 & 0.184 & 0.220 & 0.171 & 0.213 \\
    & 192 & 0.225 & 0.259 & 0.238 & 0.267 & 0.237 & 0.267 & 0.220 & 0.257 & 0.218 & 0.255 & 0.215 & 0.253 & 0.222 & 0.257 & 0.223 & 0.260 & 0.222 & 0.256 & 0.217 & 0.255 & 0.217 & 0.253 \\
    & 336 & 0.278 & 0.297 & 0.292 & 0.306 & 0.292 & 0.305 & 0.276 & 0.297 & 0.273 & 0.295 & 0.272 & 0.295 & 0.277 & 0.296 & 0.279 & 0.301 & 0.278 & 0.299 & 0.273 & 0.296 & 0.272 & 0.293 \\
    & 720 & 0.354 & 0.348 & 0.364 & 0.352 & 0.361 & 0.349 & 0.352 & 0.346 & 0.350 & 0.344 & 0.349 & 0.344 & 0.353 & 0.346 & 0.355 & 0.350 & 0.353 & 0.346 & 0.348 & 0.344 & 0.348 & 0.343 \\
    \cmidrule(lr){2-24}
    & Avg & 0.259 & 0.281 & 0.272 & 0.289 & 0.270 & 0.288 & 0.256 & 0.279 & 0.253 & 0.277 & 0.252 & 0.276 & 0.257 & 0.279 & 0.259 & 0.283 & 0.257 & 0.280 & 0.256 & 0.279 & \textbf{0.252} & \textbf{0.276} \\
    \midrule
    \multirow{5}{*}{\rotatebox{90}{Exchange}} 
    &  96 & 0.084 & 0.201 & 0.085 & 0.197 & 0.086 & 0.203 & 0.084 & 0.201 & 0.090 & 0.208 & 0.096 & 0.220 & 0.083 & 0.201 & 0.098 & 0.218 & 0.083 & 0.200 & 0.083 & 0.202 & 0.084 & 0.203 \\
    & 192 & 0.187 & 0.307 & 0.187 & 0.308 & 0.180 & 0.301 & 0.185 & 0.306 & 0.179 & 0.301 & 0.190 & 0.313 & 0.173 & 0.296 & 0.187 & 0.308 & 0.186 & 0.307 & 0.182 & 0.303 & 0.176 & 0.300 \\
    & 336 & 0.337 & 0.422 & 0.332 & 0.422 & 0.329 & 0.416 & 0.328 & 0.415 & 0.332 & 0.416 & 0.409 & 0.455 & 0.332 & 0.418 & 0.330 & 0.418 & 0.327 & 0.415 & 0.346 & 0.427 & 0.310 & 0.404 \\
    & 720 & 0.858 & 0.695 & 0.867 & 0.694 & 0.851 & 0.694 & 0.856 & 0.696 & 0.854 & 0.698 & 1.035 & 0.749 & 0.860 & 0.698 & 0.925 & 0.731 & 0.882 & 0.708 & 0.831 & 0.689 & 0.842 & 0.690 \\
    \cmidrule(lr){2-24}
    & Avg & 0.367 & 0.406 & 0.368 & 0.405 & 0.362 & 0.404 & 0.363 & 0.405 & 0.364 & 0.406 & 0.433 & 0.434 & 0.362 & 0.403 & 0.385 & 0.419  & 0.370 & 0.408 & 0.361 & 0.405 & \textbf{0.353} & \textbf{0.399} \\
    \midrule
    \multirow{5}{*}{\rotatebox{90}{ECL}} 
    &  96 & 0.193 & 0.291 & 0.190 & 0.287 & 0.190 & 0.277 & 0.175 & 0.268 & 0.178 & 0.269 & 0.183 & 0.275 & 0.171 & 0.263 & 0.171 & 0.267 & 0.181 & 0.271 & 0.164 & 0.255 & 0.164 & 0.245 \\
    & 192 & 0.199 & 0.297 & 0.204 & 0.290 & 0.194 & 0.283 & 0.183 & 0.275 & 0.185 & 0.275 & 0.190 & 0.281 & 0.188 & 0.277 & 0.181 & 0.276 & 0.197 & 0.277 & 0.178 & 0.268 & 0.173 & 0.256 \\
    & 336 & 0.216 & 0.312 & 0.227 & 0.300 & 0.211 & 0.299 & 0.199 & 0.292 & 0.202 & 0.292 & 0.205 & 0.296 & 0.205 & 0.291 & 0.197 & 0.291 & 0.200 & 0.293 & 0.190 & 0.280 & 0.189 & 0.275 \\
    & 720 & 0.257 & 0.345 & 0.257 & 0.347 & 0.254 & 0.334 & 0.240 & 0.324 & 0.245 & 0.326 & 0.248 & 0.330 & 0.244 & 0.322 & 0.237 & 0.325 & 0.205 & 0.326 & 0.235 & 0.318 & 0.229 & 0.310 \\
    \cmidrule(lr){2-24}
    & Avg & 0.216 & 0.311 & 0.220 & 0.306 & 0.212 & 0.298 & 0.199 & 0.290 & 0.203 & 0.291 & 0.207 & 0.296 & 0.202 & 0.288 & 0.197 & 0.290 & 0.196 & 0.292 & 0.192 & 0.280 & \textbf{0.189} & \textbf{0.272} \\
    \midrule
    \multirow{5}{*}{\rotatebox{90}{Traffic}} 
    &  96 & 0.472 & 0.305 & 0.449 & 0.487 & 0.483 & 0.309 & 0.309 & 0.291 & 0.458 & 0.294 & 0.506 & 0.330 & 0.465 & 0.301 & 0.478 & 0.292 & 0.463 & 0.295 & 0.442 & 0.285 & 0.429 & 0.279 \\
    & 192 & 0.474 & 0.304 & 0.505 & 0.315 & 0.495 & 0.311 & 0.457 & 0.293 & 0.465 & 0.297 & 0.503 & 0.326 & 0.470 & 0.311 & 0.469 & 0.316 & 0.470 & 0.299 & 0.452 & 0.305 & 0.442 & 0.282 \\
    & 336 & 0.491 & 0.331 & 0.514 & 0.343 & 0.504 & 0.333 & 0.474 & 0.301 & 0.480 & 0.304 & 0.517 & 0.332 & 0.498 & 0.320 & 0.482 & 0.323 & 0.486 & 0.321 & 0.473 & 0.322 & 0.456 & 0.288 \\
    & 720 & 0.523 & 0.327 & 0.511 & 0.351 & 0.521 & 0.341 & 0.509 & 0.319 & 0.515 & 0.321 & 0.552 & 0.349 & 0.514 & 0.326 & 0.516 & 0.327 & 0.504 & 0.337 & 0.497 & 0.331 & 0.486 & 0.307 \\
    \cmidrule(lr){2-24}
    & Avg & 0.490 & 0.317 & 0.504 & 0.330 & 0.501 & 0.324 & 0.472 & 0.301 & 0.480 & 0.304 & 0.520 & 0.334 & 0.487 & 0.315 & 0.486 & 0.315 & 0.481 & 0.313 & 0.466 & 0.311 & \textbf{0.453} & \textbf{0.289} \\
    \bottomrule
  \end{tabular}}
  \end{sc}
  \end{small}
  }
\end{table*}

\begin{table*}[h]
  \caption{Full results for the in-domain setting of forecasting using iTransformer. Pre-training and fine-tuning are performed on the same datasets. The standard deviations are within 0.005 for MSE and within 0.004 for MAE.}
  \label{tab:iTransformer_forecasting_indomain_full}
  \vspace{-5pt}
  \vskip 0.15in
  \centering
  \resizebox{1.02\textwidth}{!}{
  \begin{small}
  \begin{sc}
  \renewcommand{\multirowsetup}{\centering}
  \setlength{\tabcolsep}{2.0pt}
  \renewcommand\arraystretch{1.0}
  \begin{tabular}{c|c|cc|cccccccccccccc|cc}
    \toprule
    \multicolumn{2}{c}{Methods}  & \multicolumn{2}{c}{Random init.} & \multicolumn{2}{c}{TS2Vec} & \multicolumn{2}{c}{CoST} & \multicolumn{2}{c}{LaST} & \multicolumn{2}{c}{TFC} & \multicolumn{2}{c}{TST} & \multicolumn{2}{c}{Ti-MAE} & \multicolumn{2}{c}{SimMTM} & \multicolumn{2}{c}{{\textbf{TimeSiame}}}\\
    \multicolumn{2}{c}{Metric} & mse & mae & mse & mae & mse & mae & mse & mae & mse & mae & mse & mae & mse & mae & mse & mae & mse & mae \\
    \toprule
    \multirow{5}{*}{\rotatebox{90}{ETTh1}}
    &  96 & 0.386 & 0.405 & 0.404 & 0.417 & 0.404 & 0.416 & 0.397 & 0.414 & 0.384 & 0.402 & 0.383 & 0.402 & 0.384 & 0.400 & 0.382 & 0.397 & 0.378 & 0.399 \\
    & 192 & 0.441 & 0.436 & 0.459 & 0.448 & 0.457 & 0.447 & 0.449 & 0.444 & 0.439 & 0.432 & 0.436 & 0.433 & 0.436 & 0.430 & 0.435 & 0.427 & 0.429 & 0.428 \\
    & 336 & 0.487 & 0.458 & 0.502 & 0.427 & 0.503 & 0.472 & 0.493 & 0.468 & 0.482 & 0.454 & 0.478 & 0.456 & 0.481 & 0.456 & 0.477 & 0.450 & 0.471 & 0.451 \\
    & 720 & 0.503 & 0.491 & 0.531 & 0.509 & 0.523 & 0.502 & 0.520 & 0.501 & 0.496 & 0.484 & 0.492 & 0.489 & 0.492 & 0.486 & 0.485 & 0.476 & 0.483 & 0.481 \\
    \cmidrule(lr){2-20}
    & Avg & 0.454 & 0.447 & 0.474 & 0.462 & 0.472 & 0.459 & 0.465 & 0.457 & 0.450 & 0.443 & 0.447 & 0.445  & 0.448 & 0.443 & 0.445 & \textbf{0.438} & \textbf{0.440} & 0.440 \\
    \midrule
    \multirow{5}{*}{\rotatebox{90}{ETTh2}}  
    &  96 & 0.297 & 0.349 & 0.299 & 0.350 & 0.298 & 0.348 & 0.303 & 0.354 & 0.301 & 0.351 & 0.293 & 0.347 & 0.298 & 0.349 & 0.295 & 0.348 & 0.289 & 0.342 \\
    & 192 & 0.380 & 0.400 & 0.384 & 0.403 & 0.382 & 0.400 & 0.381 & 0.402 & 0.377 & 0.396 & 0.375 & 0.398 & 0.378 & 0.397 & 0.374 & 0.397 & 0.367 & 0.390 \\
    & 336 & 0.428 & 0.432 & 0.415 & 0.428 & 0.425 & 0.436 & 0.423 & 0.434 & 0.418 & 0.430 & 0.413 & 0.430 & 0.414 & 0.429 & 0.413 & 0.430 & 0.408 & 0.422 \\
    & 720 & 0.427 & 0.445 & 0.419 & 0.442 & 0.440 & 0.454 & 0.438 & 0.451 & 0.420 & 0.441 & 0.421 & 0.443 & 0.420 & 0.441 & 0.420 & 0.443 & 0.418 & 0.440 \\
    \cmidrule(lr){2-20}
    & Avg & 0.383 & 0.407 & 0.379 & 0.406 & 0.386 & 0.410 & 0.386 & 0.410 & 0.379 & 0.405 & 0.376 & 0.405 & 0.378 & 0.404 & 0.376 & 0.405 & \textbf{0.371} & \textbf{0.398} \\
    \midrule
    \multirow{5}{*}{\rotatebox{90}{ETTm1}} 
    &  96 & 0.334 & 0.368 & 0.349 & 0.379 & 0.348 & 0.378 & 0.333 & 0.369 & 0.335 & 0.368 & 0.334 & 0.373 & 0.339 & 0.374 & 0.327 & 0.364 & 0.329 & 0.366 \\
    & 192 & 0.377 & 0.391 & 0.387 & 0.396 & 0.388 & 0.397 & 0.377 & 0.392 & 0.378 & 0.390 & 0.377 & 00.396 & 0.375 & 0.391 & 0.372 & 0.386 & 0.368 & 0.386 \\
    & 336 & 0.426 & 0.420 & 0.422 & 0.418 & 0.422 & 0.418 & 0.413 & 0.414 & 0.413 & 0.412 & 0.410 & 0.417 & 0.410 & 0.412 & 0.410 & 0.410 & 0.403 & 0.408 \\
    & 720 & 0.491 & 0.459 & 0.485 & 0.454 & 0.487 & 0.455 & 0.477 & 0.451 & 0.485 & 0.452 & 0.475 & 0.455 & 0.471 & 0.446 & 0.478 & 0.450 & 0.466 & 0.445 \\
    \cmidrule(lr){2-20}
    & Avg & 0.407 & 0.410 & 0.411 & 0.412 & 0.411 & 0.412 & 0.400 & 0.407 & 0.403 & 0.406 & 0.399 & 0.410 & 0.399 & 0.406 & 0.397 & 0.403 & \textbf{0.392} & \textbf{0.401} \\
    \midrule
    \multirow{5}{*}{\rotatebox{90}{ETTm2}} 
    &  96 & 0.180 & 0.264 & 0.186 & 0.272 & 0.191 & 0.273 & 0.184 & 0.268 & 0.180 & 0.264 & 0.186 & 0.270 & 0.182 & 0.265 & 0.180 & 0.264 & 0.179 & 0.263 \\
    & 192 & 0.250 & 0.309 & 0.251 & 0.312 & 0.254 & 0.312 & 0.268 & 0.317 & 0.251 & 0.308 & 0.252 & 0.312 & 0.248 & 0.308 & 0.246 & 0.307 & 0.245 & 0.306 \\
    & 336 & 0.311 & 0.348 & 0.313 & 0.351 & 0.315 & 0.350 & 0.327 & 0.359 & 0.320 & 0.352 & 0.313 & 0.350 & 0.310 & 0.348 & 0.307 & 0.347 & 0.306 & 0.345 \\
    & 720 & 0.412 & 0.407 & 0.409 & 0.403 & 0.416 & 0.407 & 0.427 & 0.414 & 0.416 & 0.407 & 0.411 & 0.407 & 0.415 & 0.407 & 0.409 & 0.405 & 0.405 & 0.401 \\
    \cmidrule(lr){2-20}
    & Avg & 0.288 & 0.332 & 0.290 & 0.335 & 0.294 & 0.336 & 0.302 & 0.340 & 0.292 & 0.333 & 0.291 & 0.335 & 0.289 & 0.332 & 0.286 & 0.331 & \textbf{0.284} & \textbf{0.329} \\
    \midrule
    \multirow{5}{*}{\rotatebox{90}{Weather}} 
    &  96 & 0.174 & 0.214 & 0.183 & 0.226 & 0.189 & 0.231 & 0.180 & 0.223 & 0.173 & 0.213 & 0.179 & 0.221 & 0.173 & 0.212 & 0.173 & 0.212 & 0.174 & 0.217 \\
    & 192 & 0.221 & 0.254 & 0.231 & 0.266 & 0.236 & 0.269 & 0.228 & 0.262 & 0.222 & 0.257 & 0.225 & 0.261 & 0.221 & 0.256 & 0.225 & 0.258 & 0.222 & 0.256 \\
    & 336 & 0.278 & 0.296 & 0.284 & 0.303 & 0.288 & 0.306 & 0.282 & 0.301 & 0.289 & 0.298 & 0.282 & 0.303 & 0.278 & 0.297 & 0.278 & 0.298 & 0.275 & 0.295 \\
    & 720 & 0.358 & 0.349 & 0.359 & 0.351 & 0.363 & 0.354 & 0.358 & 0.350 & 0.377 & 0.350 & 0.358 & 0.352 & 0.357 & 0.349 & 0.359 & 0.349 & 0.350 & 0.346 \\
    \cmidrule(lr){2-20}
    & Avg & 0.258 & \textbf{0.278} & 0.264 & 0.287 & 0.269 & 0.290 & 0.262 & 0.284 & 0.265 & 0.280 & 0.261 & 0.284 & 0.257 & 0.279 & 0.259 & 0.279 & \textbf{0.255} & 0.279 \\
    \midrule
    \multirow{5}{*}{\rotatebox{90}{Exchange}} 
    &  96 & 0.086 & 0.206 & 0.088 & 0.208 & 0.090 & 0.213 & 0.091 & 0.212 & 0.087 & 0.208 & 0.090 & 0.211 & 0.087 & 0.208 & 0.087 & 0.211 & 0.092 & 0.215 \\
    & 192 & 0.177 & 0.299 & 0.180 & 0.302 & 0.183 & 0.306 & 0.187 & 0.310 & 0.176 & 0.300 & 0.182 & 0.305 & 0.180 & 0.303 & 0.182 & 0.304 & 0.182 & 0.306 \\
    & 336 & 0.331 & 0.417 & 0.332 & 0.418 & 0.335 & 0.420 & 0.333 & 0.421 & 0.347 & 0.428 & 0.332 & 0.419 & 0.333 & 0.418 & 0.330 & 0.411 & 0.341 & 0.426 \\
    & 720 & 0.847 & 0.691 & 0.854 & 0.697 & 0.856 & 0.699 & 0.933 & 0.736 & 0.877 & 0.709 & 0.849 & 0.699 & 0.865 & 0.703 & 0.833 & 0.669 & 0.805 & 0.679 \\
    \cmidrule(lr){2-20}
    & Avg & 0.360 & 0.403 & 0.364 & 0.406 & 0.366 & 0.410 & 0.386 & 0.420 & 0.372 & 0.411 & 0.363 & 0.409 & 0.366 & 0.408 & 0.358 & \textbf{0.399} & \textbf{0.355} & 0.407 \\
    \midrule
    \multirow{5}{*}{\rotatebox{90}{ECL}} 
    &  96 & 0.148 & 0.240 & 0.214 & 0.310 & 0.225 & 0.318 & 0.202 & 0.296 & 0.191 & 0.278 & 0.196 & 0.292 & 0.185 & 0.281 & 0.145 & 0.236 & 0.147 & 0.239 \\
    & 192 & 0.162 & 0.253 & 0.228 & 0.324 & 0.234 & 0.328 & 0.217 & 0.312 & 0.202 & 0.291 & 0.208 & 0.304 & 0.197 & 0.293 & 0.169 & 0.259 & 0.162 & 0.253 \\
    & 336 & 0.178 & 0.269 & 0.247 & 0.340 & 0.253 & 0.344 & 0.239 & 0.331 & 0.222 & 0.310 & 0.230 & 0.323 & 0.219 & 0.312 & 0.176 & 0.267 & 0.175 & 0.269 \\
    & 720 & 0.225 & 0.317 & 0.294 & 0.375 & 0.297 & 0.376 & 0.288 & 0.368 & 0.267 & 0.346 & 0.276 & 0.358 & 0.265 & 0.347 & 0.225 & 0.310 & 0.215 & 0.304 \\
    \cmidrule(lr){2-20}
    & Avg & 0.178 & 0.270 & 0.246 & 0.337 & 0.252 & 0.342 & 0.237 & 0.327 & 0.222 & 0.306 & 0.228 & 0.319 & 0.217 & 0.308 & 0.179 & 0.268 & \textbf{0.175} & \textbf{0.266} \\
    \midrule
    \multirow{5}{*}{\rotatebox{90}{Traffic}} 
    &  96 & 0.395 & 0.268 & 0.450 & 0.313 & 0.504 & 0.352 & 0.439 & 0.304 & 0.389 & 0.286 & 0.394 & 0.282 & 0.398 & 0.280 & 0.400 & 0.273 & 0.386 & 0.262 \\
    & 192 & 0.417 & 0.276 & 0.469 & 0.321 & 0.509 & 0.352 & 0.462 & 0.315 & 0.398 & 0.297 & 0.403 & 0.301 & 0.405 & 0.294 & 0.412 & 0.280 & 0.411 & 0.272 \\
    & 336 & 0.433 & 0.283 & 0.491 & 0.331 & 0.529 & 0.362 & 0.484 & 0.325 & 0.435 & 0.314 & 0.440 & 0.310 & 0.433 & 0.304 & 0.426 & 0.288 & 0.425 & 0.278 \\
    & 720 & 0.467 & 0.302 & 0.531 & 0.352 & 0.572 & 0.383 & 0.523 & 0.347 & 0.504 & 0.344 & 0.514 & 0.343 & 0.483 & 0.333 & 0.466 & 0.307 & 0.458 & 0.297 \\
    \cmidrule(lr){2-20}
    & Avg & 0.428 & 0.282 & 0.485 & 0.329 & 0.529 & 0.362 & 0.477 & 0.323 & 0.432 & 0.310 & 0.438 & 0.309 & 0.430 & 0.303 & 0.426 & 0.287 & \textbf{0.420} & \textbf{0.277} \\
    \bottomrule
  \end{tabular}
  \end{sc}
  \end{small}
  }
\end{table*}

\begin{table*}[h]
  \caption{Full results for the cross-domain setting of forecasting using PatchTST. Pre-training on the TSLD-1G dataset and fine-tune it on various target dataset. The standard deviations are within 0.005 for MSE and within 0.004 for MAE.}
  \label{tab:PatchTST_forecasting_crossdomain_full}
  \vspace{-5pt}
  \vskip 0.15in
  \centering
  \begin{small}
  \begin{sc}
  \renewcommand{\multirowsetup}{\centering}
  \setlength{\tabcolsep}{4.5pt}
  \renewcommand\arraystretch{1.0}
  \begin{tabular}{c|c|cc|cccccccccc|cc}
    \toprule
    \multicolumn{2}{c}{Methods}  & \multicolumn{2}{c}{Random init.} & \multicolumn{2}{c}{TS2Vec} & \multicolumn{2}{c}{TFC} & \multicolumn{2}{c}{TST} & \multicolumn{2}{c}{Ti-MAE} & \multicolumn{2}{c}{SimMTM} & \multicolumn{2}{c}{{\textbf{TimeSiame}}}\\
    \multicolumn{2}{c}{Metric} & mse & mae & mse & mae & mse & mae & mse & mae & mse & mae & mse & mae & mse & mae \\
    \toprule
    \multirow{5}{*}{\rotatebox{90}{ETTh1}}
    &  96 & 0.420 & 0.423 & 0.384 & 0.407 & 0.384 & 0.405 & 0.391 & 0.412 & 0.384 & 0.405 & 0.376 & 0.402 & 0.371 & 0.398 \\
    & 192 & 0.465 & 0.449 & 0.436 & 0.439 & 0.426 & 0.432 & 0.433 & 0.436 & 0.437 & 0.437 & 0.421 & 0.432 & 0.417 & 0.427 \\
    & 336 & 0.504 & 0.470 & 0.462 & 0.456 & 0.463 & 0.453 & 0.458 & 0.450 & 0.462 & 0.454 & 0.454 & 0.455 & 0.448 & 0.447 \\
    & 720 & 0.502 & 0.492 & 0.481 & 0.483 & 0.473 & 0.476 & 0.455 & 0.469 & 0.458 & 0.469 & 0.466 & 0.479 & 0.463 & 0.473 \\
    \cmidrule(lr){2-16}
    & Avg & 0.473 & 0.458 & 0.441 & 0.446 & 0.437 & 0.442 & 0.434 & 0.442 & 0.435 & 0.441 & 0.429 & 0.442 & \textbf{0.425} & \textbf{0.436} \\
    \midrule
    \multirow{5}{*}{\rotatebox{90}{ETTh2}}  
    &  96 & 0.297 & 0.345 & 0.295 & 0.347 & 0.296 & 0.347 & 0.298 & 0.346 & 0.287 & 0..342 & 0.293 & 0.347 & 0.292 & 0.345 \\
    & 192 & 0.388 & 0.400 & 0.366 & 0.392 & 0.368 & 0.392 & 0.383 & 0.401 & 0.367 & 0.391 & 0.387 & 0.409 & 0.370 & 0.394 \\
    & 336 & 0.426 & 0.434 & 0.413 & 0.426 & 0.421 & 0.433 & 0.428 & 0.435 & 0.409 & 0.425 & 0.421 & 0.428 & 0.410 & 0.427 \\
    & 720 & 0.431 & 0.446 & 0.427 & 0.450 & 0.426 & 0.445 & 0.427 & 0.446 & 0.424 & 0.444 & 0.418 & 0.443 & 0.423 & 0.444 \\
    \cmidrule(lr){2-16}
    & Avg & 0.385 & 0.406 & 0.375 & 0.404 & 0.378 & 0.404 & 0.384 & 0.407 & \textbf{0.374} & \textbf{0.403} & 0.380 & 0.407 & \textbf{0.374} & \textbf{0.403} \\
    \midrule
    \multirow{5}{*}{\rotatebox{90}{ETTm1}} 
    &  96 & 0.330 & 0.368 & 0.320 & 0.360 & 0.325 & 0.363 & 0.326 & 0.364 & 0.319 & 0.360 & 0.316 & 0.357 & 0.309 & 0.352 \\
    & 192 & 0.369 & 0.385 & 0.359 & 0.382 & 0.366 & 0.385 & 0.369 & 0.385 & 0.358 & 0.383 & 0.355 & 0.380 & 0.350 & 0.378 \\
    & 336 & 0.400 & 0.407 & 0.395 & 0.407 & 0.398 & 0.408 & 0.399 & 0.407 & 0.398 & 0.411 & 0.386 & 0.400 & 0.383 & 0.402 \\
    & 720 & 0.460 & 0.439 & 0.446 & 0.435 & 0.455 & 0.440 & 0.454 & 0.438 & 0.398 & 0.411 & 0.443 & 0.436 & 0.442 & 0.437 \\
    \cmidrule(lr){2-16}
    & Avg & 0.390 & 0.400 & 0.380 & 0.396 & 0.386 & 0.399 & 0.387 & 0.399 & 0.380 & 0.398 & 0.375 & 0.393 & \textbf{0.371} & \textbf{0.392} \\
    \midrule
    \multirow{5}{*}{\rotatebox{90}{ETTm2}} 
    &  96 & 0.175 & 0.258 & 0.176 & 0.261 & 0.176 & 0.259 & 0.188 & 0.271 & 0.179 & 0.265 & 0.177 & 0.264 & 0.182 & 0.268 \\
    & 192 & 0.247 & 0.307 & 0.245 & 0.306 & 0.243 & 0.303 & 0.258 & 0.318 & 0.256 & 0.316 & 0.247 & 0.308 & 0.243 & 0.311 \\
    & 336 & 0.309 & 0.345 & 0.310 & 0.347 & 0.303 & 0.342 & 0.334 & 0.361 & 0.325 & 0.359 & 0.309 & 0.348 & 0.314 & 0.351 \\
    & 720 & 0.408 & 0.403 & 0.432 & 0.419 & 0.407 & 0.403 & 0.431 & 0.417 & 0.415 & 0.406 & 0.416 & 0.412 & 0.406 & 0.405 \\
    \cmidrule(lr){2-16}
    & Avg & 0.285 & 0.328 & 0.291 & 0.333 & \textbf{0.282} & \textbf{0.327} & 0.303 & 0.342 & 0.294 & 0.337 & 0.287 & 0.333 & 0.286 & 0.334 \\
    \midrule
    \multirow{5}{*}{\rotatebox{90}{Weather}} 
    &  96 & 0.177 & 0.218 & 0.174 & 0.216 & 0.184 & 0.222 & 0.188 & 0.231 & 0.175 & 0.216 & 0.170 & 0.214 & 0.170 & 0.214 \\
    & 192 & 0.225 & 0.259 & 0.220 & 0.256 & 0.229 & 0.261 & 0.229 & 0.266 & 0.220 & 0.256 & 0.217 & 0.254 & 0.217 & 0.255 \\
    & 336 & 0.278 & 0.297 & 0.277 & 0.297 & 0.284 & 0.300 & 0.281 & 0.303 & 0.276 & 0.296 & 0.273 & 0.295 & 0.270 & 0.295 \\
    & 720 & 0.354 & 0.348 & 0.352 & 0.346 & 0.360 & 0.348 & 0.355 & 0.350 & 0.351 & 0.345 & 0.349 & 0.344 & 0.348 & 0.345\\
    \cmidrule(lr){2-16}
    & Avg & 0.259 & 0.281 & 0.256 & 0.279 & 0.264 & 0.283 & 0.263 & 0.288 & 0.256 & 0.278 & 0.252 & \textbf{0.277} & \textbf{0.251} & \textbf{0.277} \\
    \midrule
    \multirow{5}{*}{\rotatebox{90}{Exchange}} 
    &  96 & 0.084 & 0.201 & 0.083 & 0.201 & 0.082 & 0.200 & 0.084 & 0.202 & 0.085 & 0.203 & 0.090 & 0.209 & 0.086 & 0.204 \\
    & 192 & 0.187 & 0.307 & 0.176 & 0.298 & 0.174 & 0.297 & 0.178 & 0.299 & 0.179 & 0.302 & 0.171 & 0.297 & 0.179 & 0.301 \\
    & 336 & 0.337 & 0.422 & 0.332 & 0.416 & 0.330 & 0.416 & 0.332 & 0.417 & 0.331 & 0.417 & 0.335 & 0.419 & 0.329 & 0.416 \\
    & 720 & 0.858 & 0.695 & 0.867 & 0.700 & 0.853 & 0.696 & 0.867 & 0.698 & 0.851 & 0.696 & 0.862 & 0.690 & 0.849 & 0.694 \\
    \cmidrule(lr){2-16}
    & Avg & 0.367 & 0.406 & 0.365 & 0.404 & \textbf{0.360} & \textbf{0.402} & 0.365 & 0.404 & 0.362 & 0.405 & 0.365 & 0.404 & \textbf{0.360} & 0.404 \\
    \midrule
    \multirow{5}{*}{\rotatebox{90}{ECL}} 
    &  96 & 0.193 & 0.291 & 0.194 & 0.285 & 0.196 & 0.285 & 0.200 & 0.289 & 0.195 & 0.288 & 0.189 & 0.283 & 0.162 & 0.249 \\
    & 192 & 0.199 & 0.297 & 0.199 & 0.291 & 0.200 & 0.291 & 0.202 & 0.292 & 0.201 & 0.294 & 0.196 & 0.289 & 0.172 & 0.259 \\
    & 336 & 0.216 & 0.312 & 0.216 & 0.307 & 0.216 & 0.306 & 0.218 & 0.307 & 0.217 & 0.309 & 0.212 & 0.304 & 0.189 & 0.276 \\
    & 720 & 0.257 & 0.345 & 0.257 & 0.339 & 0.258 & 0.339 & 0.259 & 0.338 & 0.258 & 0.341 & 0.254 & 0.337 & 0.227 & 0.309 \\
    \cmidrule(lr){2-16}
    & Avg & 0.216 & 0.331 & 0.257 & 0.339 & 0.218 & 0.305 & 0.220 & 0.307 & 0.218 & 0.308 & 0.213 & 0.303 & \textbf{0.188} & \textbf{0.273} \\
    \midrule
    \multirow{5}{*}{\rotatebox{90}{Traffic}} 
    &  96 & 0.472 & 0.305 & 0.513 & 0.340 & 0.532 & 0.354 & 0.514 & 0.329 & 0.499 & 0.328 & 0.437 & 0.280 & 0.430 & 0.277 \\
    & 192 & 0.474 & 0.304 & 0.512 & 0.338 & 0.526 & 0.349 & 0.497 & 0.328 & 0.500 & 0.328 & 0.447 & 0.283 & 0.443 & 0.280 \\
    & 336 & 0.491 & 0.331 & 0.525 & 0.342 & 0.539 & 0.354 & 0.504 & 0.328 & 0.512 & 0.332 & 0.460 & 0.289 & 0.456 & 0.286 \\
    & 720 & 0.523 & 0.327 & 0.560 & 0.359 & 0.576 & 0.372 & 0.540 & 0.335 & 0.547 & 0.350 & 0.492 & 0.307 & 0.488 & 0.304 \\
    \cmidrule(lr){2-16}
    & Avg & 0.490 & 0.317 & 0.528 & 0.345 & 0.543 & 0.357 & 0.514 & 0.330 & 0.515 & 0.335 & 0.459 & 0.290 & \textbf{0.454} & \textbf{0.287} \\
    \bottomrule
  \end{tabular}
  \end{sc}
  \end{small}
\end{table*}

\begin{table*}[h]
  \caption{Full results for the in-domain setting of forecasting based on PatchTST. Pre-training and linear probing on the same dataset. The standard deviations are within 0.005 for MSE and within 0.004 for MAE. Note that the \emph{Random init.} here refers to the train-from-scratch model, which will optimize the whole model. Thus, the \emph{Random init.} is in the different setting w.r.t. other pre-training methods, where the latters are from linear probing.}
  \label{tab:PatchTST_forecasting_linear_indomain_full}
  \vspace{-5pt}
  \vskip 0.15in
  \centering
  \begin{small}
  \begin{sc}
  \renewcommand{\multirowsetup}{\centering}
  \setlength{\tabcolsep}{4.5pt}
  \renewcommand\arraystretch{1.0}
  \begin{tabular}{c|c|cc|cccccccccc|cc}
    \toprule
    \multicolumn{2}{c}{Methods}  & \multicolumn{2}{c}{Random init.} & \multicolumn{2}{c}{TS2Vec} & \multicolumn{2}{c}{TFC} & \multicolumn{2}{c}{TST} & \multicolumn{2}{c}{Ti-MAE} & \multicolumn{2}{c}{SimMTM} & \multicolumn{2}{c}{{\textbf{TimeSiame}}}\\
    \multicolumn{2}{c}{Metric} & mse & mae & mse & mae & mse & mae & mse & mae & mse & mae & mse & mae & mse & mae \\
    \toprule
    \multirow{5}{*}{\rotatebox{90}{ETTh1}}
    &  96 & 0.420 & 0.423 & 0.390 & 0.399 & 0.386 & 0.402 & 0.371 & 0.396 & 0.392 & 0.403 & 0.379 & 0.400 & 0.379 & 0.401 \\
    & 192 & 0.465 & 0.449 & 0.444 & 0.431 & 0.433 & 0.429 & 0.421 & 0.424 & 0.441 & 0.431 & 0.427 & 0.428 & 0.429 & 0.432 \\
    & 336 & 0.504 & 0.470 & 0.483 & 0.451 & 0.427 & 0.470 & 0.454 & 0.443 & 0.483 & 0.453 & 0.465 & 0.447 & 0.449 & 0.460 \\
    & 720 & 0.502 & 0.492 & 0.486 & 0.477 & 0.470 & 0.472 & 0.474 & 0.462 & 0.484 & 0.477 & 0.460 & 0.462 & 0.468 & 0.460 \\
    \cmidrule(lr){2-16}
    & Avg & 0.473 & 0.458 & 0.451 & 0.440 & 0.440 & 0.443 & \textbf{0.430} & \textbf{0.431} & 0.450 & 0.441 & 0.433 & 0.434 & 0.431 & 0.438 \\
    \midrule
    \multirow{5}{*}{\rotatebox{90}{ETTh2}}  
    &  96 & 0.297 & 0.345 & 0.290 & 0.341 & 0.291 & 0.341 & 0.287 & 0.340 & 0.324 & 0.368 & 0.299 & 0.351 & 0.281 & 0.336 \\
    & 192 & 0.388 & 0.400 & 0.371 & 0.390 & 0.369 & 0.389 & 0.389 & 0.391 & 0.405 & 0.415 & 0.377 & 0.400 & 0.362 & 0.387 \\
    & 336 & 0.426 & 0.434 & 0.428 & 0.437 & 0.418 & 0.430 & 0.411 & 0.426 & 0.442 & 0.447 & 0.420 & 0.433 & 0.406 & 0.423 \\
    & 720 & 0.431 & 0.446 & 0.430 & 0.446 & 0.423 & 0.443 & 0.421 & 0.442 & 0.463 & 0.467 & 0.422 & 0.446 & 0.418 & 0.441 \\
    \cmidrule(lr){2-16}
    & Avg & 0.385 & 0.406 & 0.380 & 0.404 & 0.375 & 0.401 & 0.372 & 0.400 & 0.409 & 0.424 & 0.380 & 0.408 & \textbf{0.367} & \textbf{0.397} \\
    \midrule
    \multirow{5}{*}{\rotatebox{90}{ETTm1}} 
    &  96 & 0.330 & 0.368 & 0.329 & 0.365 & 0.341 & 0.375 & 0.325 & 0.362 & 0.372 & 0.388 & 0.325 & 0.363 & 0.320 & 0.361  \\
    & 192 & 0.369 & 0.385 & 0.370 & 0.385 & 0.382 & 0.393 & 0.366 & 0.383 & 0.408 & 0.406 & 0.368 & 0.383 & 0.361 & 0.383 \\
    & 336 & 0.400 & 0.407 & 0.403 & 0.408 & 0.415 & 0.413 & 0.398 & 0.403 & 0.439 & 0.426 & 0.399 & 0.404 & 0.392 & 0.403 \\
    & 720 & 0.460 & 0.439 & 0.459 & 0.438 & 0.470 & 0.444 & 0.459 & 0.436 & 0.496 & 0.458 & 0.462 & 0.439 & 0.453 & 0.438 \\
    \cmidrule(lr){2-16}
    & Avg & 0.390 & 0.400 & 0.390 & 0.399 & 0.402 & 0.406 & 0.387 & 0.396 & 0.429 & 0.420 & 0.389 & 0.397 & \textbf{0.382} & \textbf{0.396} \\
    \midrule
    \multirow{5}{*}{\rotatebox{90}{ETTm2}} 
    & 96 & 0.175 & 0.258 & 0.177 & 0.264 & 0.180 & 0.268 & 0.181 & 0.269 & 0.192 & 0.276 & 0.179 & 0.267 & 0.178 & 0.265 \\
    & 192 & 0.247 & 0.307 & 0.241 & 0.304 & 0.245 & 0.309 & 0.244 & 0.308 & 0.255 & 0.314 & 0.244 & 0.307 & 0.243 & 0.305 \\
    & 336 & 0.309 & 0.345  & 0.300 & 0.342 & 0.303 & 0.346 & 0.303 & 0.345 & 0.319 & 0.354 & 0.304 & 0.346 & 0.303 & 0.343 \\
    & 720 & 0.408 & 0.403 & 0.398 & 0.398 & 0.399 & 0.401 & 0.402 & 0.400 & 0.416 & 0.406 & 0.402 & 0.401 & 0.400 & 0.399 \\
    \cmidrule(lr){2-16}
    & Avg & 0.285 & 0.328  & \textbf{0.279} & \textbf{0.327} & 0.282 & 0.331 & 0.283 & 0.331 & 0.296 & 0.338 & 0.282 & 0.330 & 0.281 & 0.328 \\
    \midrule
    \multirow{5}{*}{\rotatebox{90}{Weather}} 
    & 96 & 0.177 & 0.218  & 0.192 & 0.23 & 0.192 & 0.230 & 0.187 & 0.229 & 0.194 & 0.234 & 0.181 & 0.223 & 0.184 & 0.227 \\
    & 192 & 0.225 & 0.259 & 0.237 & 0.267 & 0.237 & 0.267 & 0.231 & 0.265 & 0.237 & 0.268 & 0.227 & 0.261 & 0.230 & 0.264 \\
    & 336 & 0.278 & 0.297 & 0.291 & 0.304 & 0.291 & 0.304 & 0.284 & 0.302 & 0.292 & 0.307 & 0.280 & 0.299 & 0.283 & 0.300 \\
    & 720 & 0.354 & 0.348 & 0.365 & 0.353 & 0.365 & 0.353 & 0.358 & 0.348 & 0.362 & 0.351 & 0.354 & 0.346 & 0.356 & 0.348 \\
    \cmidrule(lr){2-16}
    & Avg & \textbf{0.259} & \textbf{0.281} & 0.271 & 0.289 & 0.271 & 0.289 & 0.265 & 0.286 & 0.271 & 0.290 & 0.261 & 0.282 & 0.263 & 0.285 \\
    \midrule
    \multirow{5}{*}{\rotatebox{90}{Exchange}} 
    & 96 & 0.084 & 0.201  & 0.191 & 0.309 & 0.178 & 0.300 & 0.176 & 0.297 & 0.190 & 0.308 & 0.175 & 0.297 & 0.177 & 0.300 \\
    & 192 & 0.187 & 0.307 & 0.328 & 0.413 & 0.328 & 0.414 & 0.327 & 0.413 & 0.336 & 0.418 & 0.336 & 0.420 & 0.325 & 0.413 \\
    & 336 & 0.337 & 9,422 & 0.853 & 0.696 & 0.864 & 0.700 & 0.850 & 0.696 & 0.945 & 0.741 & 0.842 & 0.692 & 0.842 & 0.691 \\
    & 720 & 0.858 & 0.695 & 0.853 & 0.696 & 0.864 & 0.700 & 0.850 & 0.696 & 0.945 & 0.741 & 0.842 & 0.692 & 0.842 & 0.691 \\
    \cmidrule(lr){2-16}
    & Avg & 0.367 & 0.406 & 0.365 & 0.406 & 0.364 & 0.404 & 0.359 & \textbf{0.402} & 0.391 & 0.420 & 0.359 & 0.403 & \textbf{0.357} & 0.403 \\
    \midrule
    \multirow{5}{*}{\rotatebox{90}{ECL}} 
    & 96 & 0.193 & 0.291 & 0.239 & 0.329 & 0.218 & 0.305 & 0.183 & 0.273 & 0.211 & 0.297 & 0.180 & 0.270 & 0.177 & 0.262 \\
    & 192 & 0.199 & 0.297 & 0.238 & 0.331 & 0.218 & 0.307 & 0.187 & 0.279 & 0.212 & 0.300 & 0.177 & 0.272 & 0.183 & 0.268 \\
    & 336 & 0.216 & 0.312 & 0.254 & 0.345 & 0.232 & 0.321 & 0.203 & 0.294 & 0.227 & 0.314 & 0.201 & 0.284 & 0.198 & 0.283 \\
    & 720 & 0.257 & 0.345 & 0.295 & 0.373 & 0.273 & 0.351 & 0.244 & 0.327 & 0.269 & 0.345 & 0.243 & 0.322 & 0.239 & 0.317 \\
    \cmidrule(lr){2-16}
    & Avg & 0.216 & 0.311 & 0.257 & 0.345 & 0.235 & 0.321 & 0.204 & 0.293 & 0.230 & 0.314 & 0.200 & 0.287 & \textbf{0.199} & \textbf{0.283} \\
    \midrule
    \multirow{5}{*}{\rotatebox{90}{Traffic}} 
    & 96 & 0.472 & 0.305 & 0.778 & 0.474 & 0.703 & 0.427 & 0.600 & 0.386 & 0.724 & 0.438 & 0.550 & 0.353 & 0.491 & 0.325 \\
    & 192 & 0.474 & 0.304 & 0.730 & 0.455 & 0.651 & 0.408 & 0.573 & 0.375 & 0.677 & 0.420 & 0.563 & 0.365 & 0.486 & 0.317 \\
    & 336 & 0.491 & 0.331 & 0.741 & 0.460 & 0.659 & 0.410 & 0.583 & 0.378 & 0.685 & 0.423 & 0.587 & 0.368 & 0.498 & 0.321 \\
    & 720 & 0.523 & 0.327 & 0.780 & 0.475 & 0.699 & 0.427 & 0.619 & 0.395 & 0.724 & 0.439 & 0.602 & 0.385 & 0.532 & 0.338 \\
    \cmidrule(lr){2-16}
    & Avg & \textbf{0.490} & \textbf{0.317} & 0.757 & 0.466 & 0.678 & 0.418 & 0.594 & 0.384 & 0.703 & 0.430 & 0.576 & 0.368 & 0.502 & 0.325 \\
    \bottomrule
  \end{tabular}
  \end{sc}
  \end{small}
\end{table*}

\begin{table*}[th]
	\caption{In-domain fine-tuning results for time series classification. The model was pre-trained on datasets AD, TDBrain, and PTB, then fine-tuned on the same dataset. Accuracy (\%), Precision (\%), Recall (\%), F1 score (\%), AUROC (\%), AUPRC (\%) are recorded. We perform the experiment five times for each outcome and present the mean and standard deviation as our reported findings.}
	\label{tab:classification_indomain_full}
        \vspace{-15pt}
	\vskip 0.3in
        \setlength{\tabcolsep}{8.0pt}
        \renewcommand\arraystretch{1.1}
	\centering
        \begin{small}
        \begin{sc}
        \renewcommand{\multirowsetup}{\centering}
            \begin{tabular}{c|l|cccccc}
                \toprule
                Datasets & Methods & Accuracy & Precision & Recall & F1 Score & AUROC & AUPRC \\
                \midrule
                \multirow{10}{*}{AD} & \textcolor{gray}{Random init.} & \textcolor{gray}{$\text{80.62}_{\pm2.17}$} & \textcolor{gray}{$\text{80.51}_{\pm2.24}$} & \textcolor{gray}{$\text{80.48}_{\pm2.18}$} & \textcolor{gray}{$\text{80.48}_{\pm2.19}$} & \textcolor{gray}{$\text{86.60}_{\pm1.60}$} & \textcolor{gray}{$\text{86.48}_{\pm1.74}$} \\
                \cmidrule(lr){2-8}
                & CPC & $\text{77.40}_{\pm7.28}$ & $\text{79.91}_{\pm4.35}$ & $\text{78.52}_{\pm6.18}$ & $\text{77.09}_{\pm7.65}$ & $\text{89.81}_{\pm2.98}$ & $\text{89.49}_{\pm3.20}$ \\
                & TNC & $\text{78.58}_{\pm6.21}$ & $\text{81.10}_{\pm4.09}$ & $\text{79.97}_{\pm5.50}$ & $\text{78.43}_{\pm6.35}$ & $\text{92.26}_{\pm2.38}$ & $\text{92.10}_{\pm2.60}$ \\
                & TS2vec & $\text{81.26}_{\pm2.08}$ & $\text{81.21}_{\pm2.14}$ & $\text{81.34}_{\pm2.04}$ & $\text{81.12}_{\pm2.06}$ & $\text{89.20}_{\pm1.76}$ & $\text{88.94}_{\pm1.85}$ \\
                & CoST & $\text{73.87}_{\pm4.35}$ & $\text{77.22}_{\pm2.36}$ & $\text{75.51}_{\pm3.70}$ & $\text{73.60}_{\pm4.65}$ & $\text{89.28}_{\pm2.07}$ & $\text{88.78}_{\pm2.23}$ \\
                & LaST & $\text{72.63}_{\pm5.58}$ & $\text{75.82}_{\pm0.71}$ & $\text{73.66}_{\pm3.50}$ & $\text{72.06}_{\pm5.87}$ & $\text{84.97}_{\pm4.00}$ & $\text{84.22}_{\pm4.57}$ \\
                & TF-C & $\text{75.31}_{\pm8.27}$ & $\text{75.87}_{\pm8.73}$ & $\text{74.83}_{\pm8.98}$ & $\text{74.54}_{\pm8.85}$ & $\text{79.45}_{\pm10.23}$ & $\text{79.33}_{\pm10.57}$ \\
                & COMET  & $\text{84.50}_{\pm4.46}$ & $\text{88.31}_{\pm2.42}$ & $\text{82.95}_{\pm5.39}$ & $\text{83.33}_{\pm5.15}$ & $\text{94.44}_{\pm2.37}$ & $\text{94.43}_{\pm2.48}$ \\
                & TST & $\text{81.50}_{\pm2.16}$ & $\text{82.23}_{\pm2.12}$ & $\text{82.35}_{\pm2.16}$ & $\text{81.49}_{\pm2.16}$ & $\text{90.41}_{\pm2.06}$ & $\text{89.67}_{\pm2.42}$ \\
                & Ti-MAE & $\text{80.70}_{\pm3.73}$ & $\text{82.23}_{\pm2.92}$ & $\text{81.84}_{\pm3.39}$ & $\text{80.67}_{\pm3.73}$ & $\text{92.32}_{\pm2.80}$ & $\text{92.18}_{\pm2.93}$ \\
                & SimMTM & $\text{86.19}_{\pm1.12}$ & $\text{87.08}_{\pm1.42}$ & $\text{85.41}_{\pm1.03}$ & $\text{85.89}_{\pm1.11}$ & $\text{91.99}_{\pm0.83}$ & $\text{92.04}_{\pm0.84}$ \\
                \cmidrule(lr){2-8}
                & \textbf{TimeSiam} & $\text{\textbf{89.93}}_{\pm1.68}$ & $\text{\textbf{90.23}}_{\pm1.39}$ & $\text{\textbf{89.46}}_{\pm1.90}$ & $\text{\textbf{89.72}}_{\pm1.78}$ & $\text{\textbf{95.31}}_{\pm1.95}$ & $\text{\textbf{95.25}}_{\pm2.19}$ \\
                \midrule
                \multirow{10}{*}{TDBrain} & \textcolor{gray}{Random init.} & \textcolor{gray}{$\text{79.08}_{\pm2.33}$} & \textcolor{gray}{$\text{80.15}_{\pm2.16}$} & \textcolor{gray}{$\text{79.08}_{\pm2.33}$} & \textcolor{gray}{$\text{78.93}_{\pm2.39}$} & \textcolor{gray}{$\text{89.17}_{\pm1.94}$} & \textcolor{gray}{$\text{89.48}_{\pm1.90}$} \\
                \cmidrule(lr){2-8}
                & CPC & $\text{85.19}_{\pm2.99}$ & $\text{85.35}_{\pm2.88}$ & $\text{85.19}_{\pm2.99}$ & $\text{85.17}_{\pm3.01}$ & $\text{93.50}_{\pm2.55}$ & $\text{93.68}_{\pm2.50}$ \\
                & TNC & $\text{85.21}_{\pm1.92}$ & $\text{86.49}_{\pm1.86}$ & $\text{85.21}_{\pm1.92}$ & $\text{85.08}_{\pm1.95}$ & $\text{95.77}_{\pm1.30}$ & $\text{95.95}_{\pm1.25}$ \\
                & TS2vec & $\text{80.21}_{\pm1.69}$ & $\text{81.38}_{\pm1.97}$ & $\text{80.21}_{\pm1.69}$ & $\text{80.07}_{\pm1.69}$ & $\text{89.57}_{\pm2.31}$ & $\text{89.60}_{\pm2.37}$ \\
                & CoST & $\text{83.86}_{\pm3.71}$ & $\text{85.00}_{\pm3.00}$ & $\text{83.86}_{\pm3.71}$ & $\text{83.70}_{\pm3.89}$ & $\text{94.58}_{\pm1.90}$ & $\text{94.79}_{\pm1.79}$ \\
                & LaST & $\text{85.13}_{\pm1.85}$ & $\text{85.79}_{\pm1.54}$ & $\text{85.13}_{\pm1.85}$ & $\text{85.06}_{\pm1.90}$ & $\text{94.88}_{\pm8.26}$ & $\text{95.10}_{\pm0.81}$ \\
                & TF-C & $\text{66.62}_{\pm1.76}$ & $\text{67.15}_{\pm1.64}$ & $\text{66.62}_{\pm1.76}$ & $\text{66.35}_{\pm1.91}$ & $\text{65.43}_{\pm6.13}$ & $\text{66.18}_{\pm4.90}$ \\
                & COMET & $\text{85.47}_{\pm1.16}$ & $\text{85.68}_{\pm1.20}$ & $\text{85.47}_{\pm1.16}$ & $\text{85.45}_{\pm1.16}$ & $\text{93.73}_{\pm1.02}$ & $\text{93.96}_{\pm0.99}$ \\
                & TST & $\text{83.22}_{\pm1.91}$ & $\text{84.86}_{\pm1.08}$ & $\text{83.22}_{\pm1.91}$ & $\text{83.01}_{\pm2.03}$ & $\text{93.86}_{\pm1.10}$ & $\text{94.03}_{\pm0.99}$\\
                & Ti-MAE & $\text{88.16}_{\pm1.87}$ & $\text{88.96}_{\pm1.42}$ & $\text{88.16}_{\pm1.87}$ & $\text{88.10}_{\pm1.91}$ & $\text{\textbf{97.27}}_{\pm0.49}$ & $\text{\textbf{96.94}}_{\pm0.48}$\\
                & SimMTM & $\text{84.81}_{\pm1.54}$ & $\text{86.43}_{\pm1.07}$ & $\text{84.81}_{\pm1.54}$ & $\text{84.54}_{\pm1.67}$ & $\text{94.18}_{\pm1.57}$ & $\text{89.51}_{\pm1.52}$ \\
                \cmidrule(lr){2-8}
                & \textbf{TimeSiam} & $\text{\textbf{90.67}}_{\pm1.24}$ & $\text{\textbf{91.08}}_{\pm1.13}$ & $\text{\textbf{90.67}}_{\pm1.24}$ & $\text{\textbf{90.64}}_{\pm1.25}$ & $\text{96.96}_{\pm0.80}$ & $\text{96.82}_{\pm0.82}$ \\
                \midrule
                \multirow{10}{*}{PTB} & \textcolor{gray}{Random init.} & \textcolor{gray}{$\text{84.19}_{\pm1.29}$} & \textcolor{gray}{$\text{83.35}_{\pm1.68}$} & \textcolor{gray}{$\text{78.46}_{\pm2.50}$} & \textcolor{gray}{$\text{80.33}_{\pm2.02}$} & \textcolor{gray}{$\text{89.55}_{\pm1.83}$} & \textcolor{gray}{$\text{83.61}_{\pm2.68}$} \\
                \cmidrule(lr){2-8}
                & CPC & $\text{88.30}_{\pm3.07}$ & $\text{88.90}_{\pm1.00}$ & $\text{81.54}_{\pm6.51}$ & $\text{83.75}_{\pm5.67}$ & $\text{89.86}_{\pm3.87}$ & $\text{88.68}_{\pm2.89}$ \\
                & TNC & $\text{90.53}_{\pm2.92}$ & $\text{89.01}_{\pm2.87}$ & $\text{87.06}_{\pm5.22}$ & $\text{87.82}_{\pm4.13}$ & $\text{93.12}_{\pm2.21}$ & $\text{91.01}_{\pm1.55}$ \\
                & TS2vec & $\text{85.14}_{\pm1.66}$ & $\text{87.82}_{\pm2.21}$  & $\text{76.84}_{\pm3.99}$ & $\text{79.66}_{\pm3.63}$ & $\text{90.50}_{\pm1.59}$ & $\text{90.07}_{\pm1.73}$ \\
                & CoST & $\text{88.61}_{\pm1.36}$ & $\text{87.75}_{\pm1.23}$  & $\text{80.23}_{\pm2.39}$ & $\text{83.81}_{\pm2.33}$ & $\text{93.79}_{\pm2.36}$ & $\text{93.01}_{\pm2.37}$ \\
                & LaST & $\text{89.22}_{\pm3.10}$ & $\text{89.12}_{\pm2.71}$  & $\text{83.32}_{\pm5.54}$ & $\text{85.45}_{\pm4.66}$ & $\text{94.91}_{\pm1.13}$ & $\text{91.79}_{\pm4.25}$ \\
                & TF-C & $\text{87.50}_{\pm2.43}$ & $\text{85.50}_{\pm3.04}$ & $\text{82.68}_{\pm4.04}$ & $\text{83.77}_{\pm3.50}$ & $\text{77.59}_{\pm19.22}$ & $\text{80.62}_{\pm15.10}$ \\
                & COMET & $\text{87.84}_{\pm1.98}$ & $\text{87.67}_{\pm1.72}$ & $\text{81.14}_{\pm3.68}$ & $\text{83.45}_{\pm3.22}$ & $\text{92.95}_{\pm1.56}$ & $\text{87.47}_{\pm2.82}$ \\
                & TST & $\text{84.25}_{\pm3.29}$ & $\text{84.05}_{\pm3.95}$ & $\text{74.83}_{\pm5.49}$ & $\text{77.45}_{\pm5.59}$ & $\text{90.44}_{\pm3.05}$ & $\text{85.74}_{\pm3.25}$ \\
                & Ti-MAE & $\text{88.39}_{\pm1.78}$ & $\text{88.55}_{\pm1.51}$ & $\text{81.76}_{\pm3.13}$ & $\text{84.23}_{\pm2.70}$ & $\text{90.37}_{\pm5.70}$ & $\text{88.76}_{\pm5.00}$ \\
                & SimMTM & $\text{90.04}_{\pm1.23}$ & $\text{89.09}_{\pm1.33}$ & $\text{85.52}_{\pm2.57}$ & $\text{87.05}_{\pm2.38}$ & $\text{92.68}_{\pm1.23}$ & $\text{90.14}_{\pm3.01}$ \\
                \cmidrule(lr){2-8}
                & \textbf{TimeSiam} & $\text{\textbf{91.32}}_{\pm2.92}$ & $\text{\textbf{89.97}}_{\pm2.89}$ & $\text{\textbf{88.02}}_{\pm4.93}$ & $\text{\textbf{88.84}}_{\pm4.12}$ & $\text{\textbf{96.42}}_{\pm1.51}$ & $\text{\textbf{94.33}}_{\pm2.09}$ \\
                \bottomrule
            \end{tabular}
          \vspace{-5pt}
          \end{sc}
           \end{small}
\end{table*}

\begin{table*}[h]
	\caption{Cross-domain fine-tuning results for time series classification. The model is pre-trained on the TSLD-1G dataset and fine-tuned on AD, TDBrain, and PTB. Accuracy (\%), Precision (\%), Recall (\%), F1 score (\%), AUROC (\%), AUPRC (\%) are recorded. We perform the experiment five times for each outcome and present the mean and standard deviation as our reported findings.}
	\label{tab:classification_crossdomain_full}
        \vspace{-15pt}
	\vskip 0.3in
        \setlength{\tabcolsep}{8.0pt}
        \renewcommand\arraystretch{1.1}
	\centering
        \begin{small}
        \begin{sc}
        \renewcommand{\multirowsetup}{\centering}
            \begin{tabular}{c|l|cccccc}
                \toprule
                Datasets & Methods & Accuracy & Precision & Recall & F1 Score & AUROC & AUPRC \\
                \midrule
                \multirow{7}{*}{AD} & \textcolor{gray}{Random init.} & \textcolor{gray}{$\text{80.62}_{\pm2.17}$} & \textcolor{gray}{$\text{80.51}_{\pm2.24}$} & \textcolor{gray}{$\text{80.48}_{\pm2.18}$} & \textcolor{gray}{$\text{80.48}_{\pm2.19}$} & \textcolor{gray}{$\text{86.60}_{\pm1.60}$} & \textcolor{gray}{$\text{86.48}_{\pm1.74}$} \\
                \cmidrule(lr){2-8}
                & TS2vec & $\text{80.59}_{\pm6.45}$ & $\text{81.77}_{\pm8.72}$ & $\text{81.61}_{\pm9.20}$ & $\text{80.53}_{\pm9.55}$ & $\text{90.31}_{\pm6.38}$ & $\text{90.00}_{\pm7.94}$ \\
                & TF-C & $\text{87.98}_{\pm1.77}$ & $\text{88.30}_{\pm1.68}$ & $\text{88.30}_{\pm1.69}$ & $\text{87.90}_{\pm1.75}$ & $\text{95.56}_{\pm1.52}$ & $\text{95.43}_{\pm1.54}$ \\
                & TST & $\text{82.60}_{\pm3.71}$ & $\text{83.81}_{\pm2.63}$ & $\text{83.35}_{\pm3.02}$ & $\text{82.51}_{\pm3.66}$ & $\text{93.05}_{\pm2.17}$ & $\text{92.75}_{\pm2.50}$ \\
                & Ti-MAE & $\text{80.40}_{\pm5.26}$ & $\text{81.72}_{\pm5.02}$ & $\text{81.13}_{\pm4.66}$ & $\text{80.31}_{\pm5.18}$ & $\text{91.32}_{\pm4.48}$ & $\text{91.16}_{\pm4.75}$ \\
                & SimMTM & $\text{87.74}_{\pm1.78}$ & $\text{87.66}_{\pm1.91}$ & $\text{87.78}_{\pm1.78}$ & $\text{87.63}_{\pm1.78}$ & $\text{94.73}_{\pm1.32}$ & $\text{94.71}_{\pm1.36}$ \\
                \cmidrule(lr){2-8}
                & \textbf{TimeSiam} & $\text{\textbf{90.47}}_{\pm2.04}$ & $\text{\textbf{90.50}}_{\pm2.01}$ & $\text{\textbf{90.21}}_{\pm2.13}$ & $\text{\textbf{90.32}}_{\pm2.09}$ & $\text{\textbf{96.34}}_{\pm1.36}$ & $\text{\textbf{96.41}}_{\pm1.39}$\\
                \midrule
                \multirow{7}{*}{TDBrain} & \textcolor{gray}{Random init.} & \textcolor{gray}{$\text{79.08}_{\pm2.33}$} & \textcolor{gray}{$\text{80.15}_{\pm2.16}$} & \textcolor{gray}{$\text{79.08}_{\pm2.33}$} & \textcolor{gray}{$\text{78.93}_{\pm2.39}$} & \textcolor{gray}{$\text{89.17}_{\pm1.94}$} & \textcolor{gray}{$\text{89.48}_{\pm1.90}$} \\
                \cmidrule(lr){2-8}
                & TS2vec & $\text{85.58}_{\pm8.16}$ & $\text{86.26}_{\pm7.78}$ & $\text{85.58}_{\pm8.16}$ & $\text{85.45}_{\pm8.32}$ & $\text{94.44}_{\pm4.03}$ & $\text{94.69}_{\pm3.79}$ \\
                & TF-C & $\text{82.84}_{\pm2.57}$ & $\text{83.22}_{\pm2.58}$ & $\text{82.84}_{\pm2.57}$ & $\text{82.79}_{\pm2.58}$ & $\text{92.13}_{\pm2.17}$ & $\text{92.28}_{\pm2.11}$ \\
                & TST & $\text{83.65}_{\pm2.52}$ & $\text{84.75}_{\pm2.27}$ & $\text{83.65}_{\pm2.52}$ & $\text{83.51}_{\pm2.59}$ & $\text{93.41}_{\pm2.13}$ & $\text{93.58}_{\pm2.08}$\\
                & Ti-MAE & $\text{85.22}_{\pm2.40}$ & $\text{82.85}_{\pm2.01}$ & $\text{82.42}_{\pm2.47}$ & $\text{82.38}_{\pm2.53}$ & $\text{90.25}_{\pm1.39}$ & $\text{90.26}_{\pm1.36}$\\
                & SimMTM & $\text{85.29}_{\pm1.87}$ & $\text{86.61}_{\pm2.45}$ & $\text{\textbf{86.23}}_{\pm2.59}$ & $\text{\textbf{86.19}}_{\pm2.62}$ & $\text{94.00}_{\pm1.78}$ & $\text{93.95}_{\pm1.84}$ \\
                \cmidrule(lr){2-8}
                & \textbf{TimeSiam} & $\text{\textbf{86.26}}_{\pm2.54}$ & $\text{\textbf{87.17}}_{\pm2.07}$ & $\text{80.26}_{\pm2.54}$ & $\text{86.17}_{\pm2.62}$ & $\text{\textbf{95.41}}_{\pm1.35}$ & $\text{\textbf{95.56}}_{\pm1.30}$\\
                \midrule
                \multirow{7}{*}{PTB} & \textcolor{gray}{Random init.} & \textcolor{gray}{$\text{84.19}_{\pm1.29}$} & \textcolor{gray}{$\text{83.35}_{\pm1.68}$} & \textcolor{gray}{$\text{78.46}_{\pm2.50}$} & \textcolor{gray}{$\text{80.33}_{\pm2.02}$} & \textcolor{gray}{$\text{89.55}_{\pm1.83}$} & \textcolor{gray}{$\text{83.61}_{\pm2.68}$} \\
                \cmidrule(lr){2-8}
                & TS2vec & $\text{89.23}_{\pm7.76}$ & $\text{89.58}_{\pm7.15}$ & $\text{\textbf{89.23}}_{\pm7.76}$ & $\text{\textbf{89.17}}_{\pm7.88}$ & $\text{\textbf{96.13}}_{\pm3.82}$ & $\text{\textbf{96.27}}_{\pm3.63}$\\
                & TF-C & $\text{89.18}_{\pm1.89}$ & $\text{88.63}_{\pm2.21}$ & $\text{84.48}_{\pm3.98}$ & $\text{85.64}_{\pm2.66}$ & $\text{94.31}_{\pm1.71}$ & $\text{91.52}_{\pm2.79}$\\
                & TST & $\text{85.81}_{\pm5.92}$ & $\text{85.80}_{\pm4.96}$ & $\text{77.40}_{\pm1.02}$ & $\text{79.32}_{\pm1.07}$ & $\text{92.04}_{\pm2.89}$ & $\text{86.27}_{\pm4.97}$\\
                & Ti-MAE & $\text{86.67}_{\pm2.55}$ & $\text{85.91}_{\pm1.10}$ & $\text{80.32}_{\pm6.76}$ & $\text{81.83}_{\pm4.94}$ & $\text{92.60}_{\pm4.45}$ & $\text{91.08}_{\pm3.78}$\\
                & SimMTM & $\text{85.64}_{\pm1.68}$ & $\text{85.94}_{\pm1.58}$ & $\text{77.01}_{\pm3.00}$ & $\text{79.80}_{\pm2.78}$ & $\text{92.93}_{\pm0.68}$ & $\text{88.03}_{\pm2.31}$ \\
                \cmidrule(lr){2-8}
                & \textbf{TimeSiam} & $\text{\textbf{90.45}}_{\pm1.98}$ & $\text{\textbf{89.58}}_{\pm1.53}$ & $\text{86.11}_{\pm3.75}$ & $\text{87.53}_{\pm2.83}$ & $\text{93.13}_{\pm2.32}$ & $\text{89.94}_{\pm3.10}$ \\
                \bottomrule
            \end{tabular}
          \vspace{-5pt}
          \end{sc}
          \end{small}
\end{table*}

\begin{table*}[h]
  \caption{Ablation studies were conducted on TimeSiam. ``W/o Siamese'' refers to solely focusing on modeling subseries itself, without incorporating Siamese modeling. ``W/o Masking'' indicates the absence of mask augmentation in the current subseries.}\label{tab:forecast_in_domain_abs_full}
  \label{tab:lra}
  \vspace{-5pt}
  \vskip 0.15in
  \centering
  \begin{sc}
  \begin{small}
  \renewcommand{\multirowsetup}{\centering}
  \setlength{\tabcolsep}{12pt}
  \renewcommand\arraystretch{0.9}
  \begin{tabular}{c|c|ccccccccccccccccccc}
    \toprule
    \multicolumn{2}{c}{Input-96} & \multicolumn{2}{c}{Random Init.} & \multicolumn{2}{c}{W/o Siamese}  & \multicolumn{2}{c}{W/o Masking} & \multicolumn{2}{c}{\textbf{TimeSiam}}  \\
    \cmidrule(lr){3-4} \cmidrule(lr){5-6} \cmidrule(lr){7-8} \cmidrule(lr){9-10}
    \multicolumn{2}{c}{Predict-\emph{O}} & MSE & MAE & MSE & MAE & MSE & MAE & MSE & MAE \\
    \toprule
    \multirow{6}{*}{ETTh1}
    &  96 & 0.420 & 0.423 & 0.377 & 0.401 & 0.381 & 0.403 & 0.378 & 0.401 \\
    & 192 & 0.465 & 0.449 & 0.423 & 0.430 & 0.430 & 0.431 & 0.422 & 0.430 \\
    & 336 & 0.504 & 0.470 & 0.458 & 0.451 & 0.466 & 0.452 & 0.459 & 0.452 \\
    & 720 & 0.502 & 0.492 & 0.471 & 0.478 & 0.470 & 0.474 & 0.459 & 0.437 \\
    \cmidrule(lr){2-10} 
    & Avg & 0.473 & 0.458 & 0.432 & 0.440 & 0.437 & 0.440 & \textbf{0.429} & \textbf{0.437} \\
    \midrule
    \multirow{6}{*}{ETTh2}  
    & 96 & 0.297 & 0.345 & 0.291 & 0.346 & 0.289 & 0.339 & 0.293 & 0.345 \\
    & 192 & 0.388 & 0.400 & 0.375 & 0.396 & 0.378 & 0.393 & 0.370 & 0.392 \\
    & 336 & 0.426 & 0.434 & 0.416 & 0.432 & 0.412 & 0.426 & 0.410 & 0.424 \\
    & 720 & 0.431 & 0.446 & 0.421 & 0.446 & 0.420 & 0.441 & 0.418 & 0.440 \\
    \cmidrule(lr){2-10} 
    & Avg & 0.385 & 0.406 & 0.376 & 0.405 & 0.375 & 0.400 & \textbf{0.373} & \textbf{0.400} \\
    \midrule
    \multirow{6}{*}{ETTm1} 
    & 96 & 0.330 & 0.368 & 0.317 & 0.359 & 0.333 & 0.368 & 0.319 & 0.360 \\
    & 192 & 0.369 & 0.385 & 0.363 & 0.387 & 0.367 & 0.385 & 0.353 & 0.379 \\
    & 336 & 0.400 & 0.407 & 0.385 & 0.403 & 0.400 & 0.409 & 0.383 & 0.402 \\
    & 720 & 0.460 & 0.439 & 0.444 & 0.438 & 0.459 & 0.442 & 0.440 & 0.436 \\
    \cmidrule(lr){2-10} 
    & Avg & 0.390 & 0.400 & 0.377 & 0.397 & 0.390 & 0.410 & \textbf{0.374} & \textbf{0.394} \\
    \midrule
    \multirow{6}{*}{ETTm2} 
    & 96 & 0.175 & 0.258 & 0.177 & 0.262 & 0.177 & 0.261 & 0.175 & 0.261 \\
    & 192 & 0.247 & 0.307 & 0.241 & 0.303 & 0.243 & 0.303 & 0.241 & 0.303 \\
    & 336 & 0.309 & 0.345 & 0.302 & 0.343 & 0.307 & 0.347 & 0.300 & 0.341 \\
    & 720 & 0.408 & 0.403 & 0.398 & 0.398 & 0.405 & 0.404 & 0.399 & 0.398 \\
    \cmidrule(lr){2-10} 
    & Avg & 0.285 & 0.328 & 0.280 & 0.327 & 0.283 & 0.329 & \textbf{0.279} & \textbf{0.326} \\
    \midrule
    \multirow{6}{*}{Weather}
    &  96 & 0.177 & 0.218 & 0.174 & 0.219 & 0.176 & 0.219 & 0.171 & 0.213 \\
    & 192 & 0.225 & 0.259 & 0.221 & 0.258 & 0.224 & 0.259 & 0.217 & 0.253 \\
    & 336 & 0.278 & 0.297 & 0.275 & 0.296 & 0.279 & 0.299 & 0.272 & 0.293 \\
    & 720 & 0.354 & 0.384 & 0.353 & 0.345 & 0.356 & 0.350 & 0.348 & 0.343 \\
    \cmidrule(lr){2-10} 
    & Avg & 0.259 & 0.281 & 0.256 & 0.280 & 0.259 & 0.282 & \textbf{0.252} & \textbf{0.276} \\
    \midrule
    \multirow{6}{*}{Exchange}  
    & 96 & 0.084 & 0.201 & 0.089 & 0.209 & 0.083 & 0.201 & 0.084 & 0.203 \\
    & 192 & 0.187 & 0.307 & 0.196 & 0.314 & 0.173 & 0.297 & 0.176 & 0.300 \\
    & 336 & 0.337 & 0.422 & 0.334 & 0.419 & 0.341 & 0.424 & 0.310 & 0.404 \\
    & 720 & 0.858 & 0.695 & 0.856 & 0.700 & 0.856 & 0.698 & 0.842 & 0.690 \\
    \cmidrule(lr){2-10} 
    & Avg & 0.367 & 0.406 & 0.369 & 0.411 & 0.363 & 0.405 & \textbf{0.353} & \textbf{0.399} \\
    \midrule
    \multirow{6}{*}{ECL} 
    & 96 & 0.193 & 0.291 & 0.164 & 0.250 & 0.165 & 0.253 & 0.164 & 0.245 \\
    & 192 & 0.199 & 0.297 & 0.175 & 0.261 & 0.177 & 0.258 & 0.173 & 0.256 \\
    & 336 & 0.216 & 0.312 & 0.191 & 0.278 & 0.190 & 0.274 & 0.189 & 0.275 \\
    & 720 & 0.257 & 0.345 & 0.230 & 0.312 & 0.232 & 0.311 & 0.229 & 0.310 \\
    \cmidrule(lr){2-10} 
    & Avg & 0.216 & 0.331 & 0.190 & 0.275 & 0.191 & 0.274 & \textbf{0.189} & \textbf{0.272} \\
    \midrule
    \multirow{6}{*}{Traffic} 
    & 96 & 0.472 & 0.305 & 0.433 & 0.281 & 0.438 & 0.283 & 0.429 & 0.279 \\
    & 192 & 0.474 & 0.304 & 0.446 & 0.287 & 0.447 & 0.287 & 0.442 & 0.282 \\
    & 336 & 0.491 & 0.331 & 0.459 & 0.288 & 0.459 & 0.289 & 0.456 & 0.288 \\
    & 720 & 0.523 & 0.327 & 0.490 & 0.306 & 0.494 & 0.307 & 0.486 & 0.307 \\
    \cmidrule(lr){2-10} 
    & Avg & 0.490 & 0.317 & 0.457 & 0.291 & 0.460 & 0.292 & \textbf{0.453} & \textbf{0.289} \\
    \bottomrule
  \end{tabular} 
  \end{small}
  \end{sc}
\end{table*}

\end{document}